%% file: main.tex
\documentclass[acmtog,table,xcdraw]{acmart}
\acmSubmissionID{}
\input{macros}

\usepackage{booktabs} 

\usepackage{graphicx}
\usepackage{comment}

\usepackage{gensymb}
\usepackage[ruled]{algorithm2e} 

\SetAlFnt{\small}
\SetAlCapFnt{\small}
\SetAlCapNameFnt{\small}
\SetAlCapHSkip{0pt}

\acmJournal{TOG}

\setcopyright{none}

\begin{document}
\title{Generative Tweening: Long-term Inbetweening of 3D Human Motions}

\author{Yi Zhou}
\affiliation{%
  \institution{University of Southern California}
  \country{USA}}
\email{zhou859@usc.edu}

\author{Jingwan Lu}
\affiliation{%
  \institution{Adobe Research}
  \country{USA}}
\email{jlu@adobe.com}

\author{Connelly Barnes}
\affiliation{%
  \institution{Adobe Research}
  \country{USA}}
\email{cobarnes@adobe.com}

\author{Jimei Yang}
\affiliation{%
  \institution{Adobe Research}
  \country{USA}}
\email{jimyang@adobe.com}

\author{Sitao Xiang}
\affiliation{%
  \institution{University of Southern California}
  \country{USA}}
\email{sitaoxia@usc.edu}

\author{Hao Li}
\affiliation{%
  \institution{University of Southern California, USC Institute for Creative Technologies, Pinscreen}
  \country{USA}}
\email{hao@hao-li.com}

\begin{teaserfigure}
\centering
\includegraphics[width=0.8\textwidth]{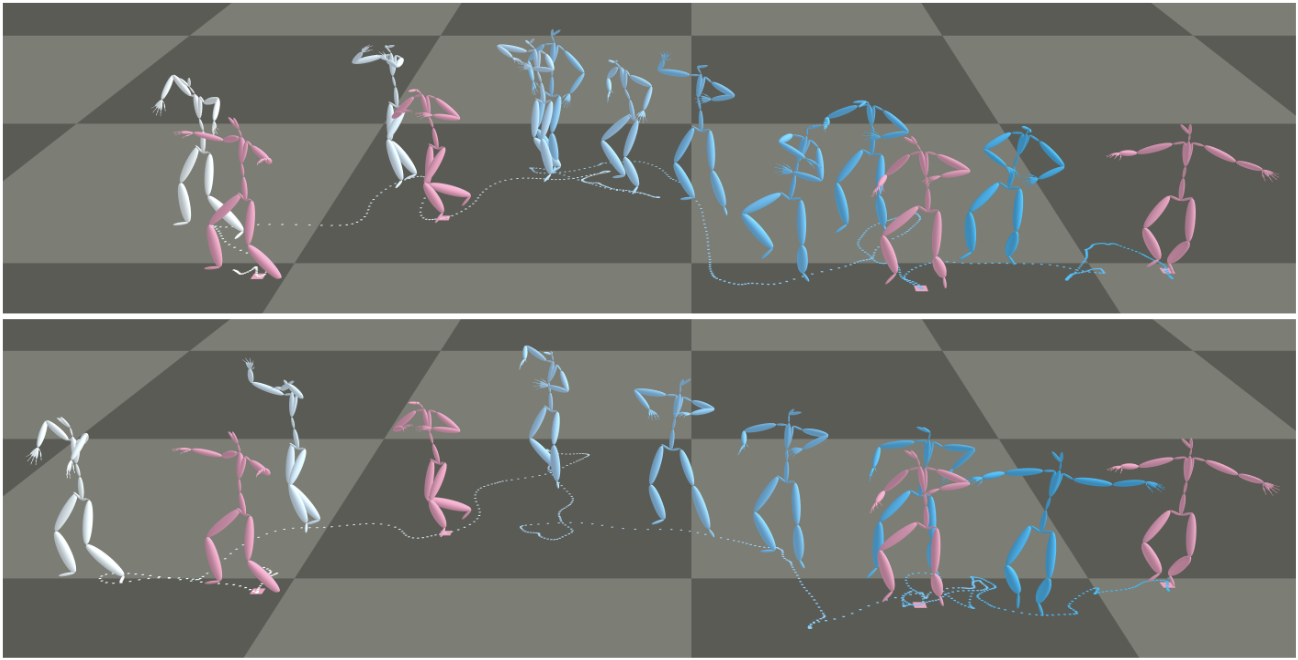}
\caption{Given the same four keyframes (pink), we show an example where our method generates two different and long motion sequences that interpolates these keyframes. Notice that the synthesized frames contain complex natural poses and meaningful variations. A subset of the synthesized frames are rendered from blue to white in the order of time. The synthesized trajectories are rendered as small dots on the ground. 
}
\label{fig:teaser}
\end{teaserfigure}

\begin{abstract}
The ability to generate complex and realistic human body animations at scale, while following specific artistic constraints, has been a fundamental goal for the game and animation industry for decades.
Popular techniques include key-framing, physics-based simulation, and database methods via motion graphs.  Recently, motion generators based on deep learning have been introduced. Although these learning models can automatically generate highly intricate stylized motions of arbitrary length, they still lack user control.
To this end, we introduce the problem of long-term inbetweening, which involves automatically  synthesizing complex motions over a long time interval given very sparse keyframes by users. We identify a number of challenges related to this problem, including maintaining biomechanical and keyframe constraints, preserving natural motions, and designing the entire motion sequence holistically while considering all constraints. We introduce a biomechanically constrained generative adversarial network that performs long-term inbetweening of human motions, conditioned on keyframe constraints. This network uses a novel two-stage approach where it first predicts local motion in the form of joint angles, and then predicts global motion, i.e. the global path that the character follows. Since there are typically a number of possible motions that could satisfy the given user constraints, we also enable our network to generate a variety of outputs with a scheme that we call Motion DNA. This approach allows the user to manipulate and influence the output content by feeding seed motions (DNA) to the network. Trained with 79 classes of captured motion data, our network performs robustly on a variety of highly complex motion styles.
\end{abstract}

%
%

%
%


\maketitle

\input{introduction_new.tex}
\input{related_work.tex}

\input{method.tex}
\input{experiments_new.tex}

\input{Limitation.tex}
\input{additional_features.tex}

\input{conclusion.tex}

\bibliographystyle{ACM-Reference-Format}
\bibliography{bibliography}

\input{appendix.tex}

\end{document}

%% file: macros.tex
\newcommand{\ignorethis} [1] {}

\newcommand{\sectnum    } [1] {\ref{#1}}

\newcommand{\sect       } [1] {Section~\sectnum{#1}}

\newcommand{\fignum     } [1] {\ref{#1}}
\newcommand{\fig        } [1] {Figure~\fignum{#1}}

\newcommand{\colornote}[3]{{\color{#1}\bf{#2: #3}\normalfont}}  

\iffalse 
        \newcommand{\cyn}[1]{}
        \newcommand{\connelly}[1]{}
        \newcommand{\jimei}[1]{}
        \newcommand{\yi}[1]{}
\else 
    \newcommand {\cyn}[1]{\colornote{green}{Cyn}{#1}}
    \newcommand {\connelly}[1]{\colornote{blue}{Connelly}{#1}}
    \newcommand {\jimei}[1]{\colornote{cyan}{Jimei}{#1}}
    \newcommand {\yi}[1]{\colornote{red}{Yi}{#1}}
\fi

%% file: introduction_new.tex
\section{Introduction}
There is a great demand for producing convincing performances for CG characters in the game and animation industry. In order to obtain realistic motion data, production studios still rely mainly on manually keyframed body animations and professional motion capture systems~\cite{kovar2008motion}.
Traditional keyframe animations require very dense pose specifications to ensure the naturalness of the motion, which is known to be a time-consuming and expensive task, requiring highly skilled animators. 

Recent deep learning approaches have shown promising results in the automatic motion synthesis for locomotion~\cite{holden2016deep,Harvey:2018:Recurrent} and more complex performances such as playing basketball~\cite{liu2018learning} and soccer~\cite{hong2019}. However, these methods typically respect high-level controls such as the path in locomotion synthesis~\cite{holden2016deep} or task-specific goals. In contrast, when producing more expressive or complicated animations, an artist may find it useful to have more precise control on specific frames~\cite{igarashi2007spatial}. 

In this paper, we aim to synthesize complex human motions of arbitrary length where keyframes are given by users at flexible space-time locations with varying densities.
%
We call this problem \emph{long-term inbetweening.} Users can iterate on their motion designs by adjusting the keyframes. 
With very sparse keyframes, our system can improvise a variety of motions under the same keyframe constraints. We show one such example in~\fig{fig:teaser}. 

The first question consists of how to design a network to generate realistic motions. The dataset contains a limited number of short motion clips and each clip performs only one type of motion. But in real application scenarios, one might need to simulate the transitions between keyframes from the same or different motion clips or even different motion categories, and such transitions are expensive to capture and thus do not exist in public mocap datasets. 

The second question consists of how to synthesize a sequence that can precisely follow the keyframes. To ensure the coherence of contents of the entire sequence, the motion within an interval should be synthesized holistically considering the global context of all the keyframes that could influence the current interval, instead of only two keyframes at each endpoint of the interval. Moreover, unlike previous works that generate motions based only on initial poses~\cite{li2017auto,yan2018mt}, long-term inbetweening requires the generator to perform sophisticated choreography to make sure the synthesized motion not only looks natural but also reaches keyframe poses at the specified times and positions. A small difference in the middle of a motion can result in big time or space displacements at keyframes. The reason for this is that the local movements of the limbs and the body affect the speed and direction of the character, and the integral of the velocities of sequential frames determine the global motion. Therefore, these differences can accumulate throughout the sequence and any modifications to a motion must consider both local and global effects. 


Facing the challenges above, we propose a conditional generative adversarial network (GAN) that can learn natural motions from a motion capture database (CMU dataset)~\cite{cmu}. We propose a two-stage method, where we first synthesize local motion, and then predict global motion. Our computation is highly efficient, where one minute of motion can be generated automatically in half a second. Real-time updates are possible for shorter sequences.

To avoid synthesizing unrealistic body poses, we embed body biomechnical constraints into the network by restricting the rotation range of local joints with a Euler angle representation. Following the observations of Zhou~et~al.~\shortcite{zhou2019}, for the local joints, we carefully choose the order and the ranges of the Euler angles to avoid discontinuities in the angular representation. 
We call the integration of this angular representation with forward kinematics Range Constrained Forward Kinematics (RC-FK) layer.

We also develop a design concept called Motion DNA to enable the network to generate a variety of motions satisfying the same keyframe constraints. The motion DNA seeds are inferred from a set of reference poses which are the representative frames of real motion sequences. By choosing the type of representative frames, the user is able to influence the style of the output motion.

Here are our main contributions:
We believe that our approach is a key component for production settings where fast turn-around is critical, such as pre-visualization or the animation of secondary characters in large scenes. Furthermore, we also anticipate potential adoption of our method in non-professional content creation settings such as educational and consumer apps. 
\begin{itemize}
\item We introduce the first deep generative model that can: 1) synthesize high-fidelity and natural motions between key frames of arbitrary lengths automatically, 2) ensure exact keyframe interpolation, 3) support motion style variations in the output, and 4) mix characteristics of multiple classes of motions. Existing methods do not support all these capabilities.
\item We propose a novel GAN architecture and training approach for 
automatic long-term motion inbetweening. Our two-stage approach makes the problem tractable by first predicting the local and then the global motion. 
\item We propose a novel Range Constrained Forward Kinematics (RC-FK) layer to embed body biomechanical constraints in the network.
\item We introduce the concept of motion DNA which allows the network to generate a variety of motion sequences from a single set of sparse user-provided keyframes.
\item Our method is significantly faster than state-of-the-art motion graph techniques, while providing intuitive control and ensuring high-quality output. We also plan to release our code and data to the public, upon acceptance of this paper.
\end{itemize}

%% file: related_work.tex
\section{Related Work}
There are many works in human motion synthesis using statistical, data-driven and physics-based methods~\cite{rose1998verbs, rose2001artist, kovar2004automated, grochow2004style, mukai2005geostatistical, LevineWHPK12, lee2002interactive, park2002line, safonova2007construction, likhachev2004ara, won2014generating, AllenF09, TanGLT14, LevineK14, MordatchLAPT15, peng2017deeploco, yu2018learning, jiang2019synthesis, liu2018learning, hong2019, naderi2017discovering, liu2017learning, peng2018deepmimic, lee2019scalable}. A comprehensive survey is beyond the scope of this work. We discuss more related papers below on transition synthesis, deep learning and biomechanic constraints.

\textbf{Transition Synthesis.} Statistical motion models have been used to generate transitions in human animations. Chai and Hodgins~\shortcite{ChaiH07} and Min~et~al.~\shortcite{MinCC09} developed MAP optimization frameworks, which can create transitions and also follow other constraints. Wang et~al.~\shortcite{WangFH08} use Gaussian process dynamical models to create transitions and synthesize motion. Lehrmann~et~al.~\shortcite{LehrmannGN14} use a nonlinear Markov model called a dynamic forest model to synthesize transitions and perform action recognition. Agrawal and van de Panne~\shortcite{agrawal2016task} develop a model for synthesizing complex locomotions including side steps, toe pivots, heel pivots, and foot slides, which allows task-specific locomotion to be produced. Harvey~and~Pal~\shortcite{Harvey:2018:Recurrent} use a recurrent network model based on long short-term memory (LSTM) to synthesize human locomotion, including for transitions. Unlike Harvey~and~Pal~\shortcite{Harvey:2018:Recurrent}, who train on a fixed interval of 60 frames for locomotion, our method can synthesize longer-term motions across different kinds of movement (e.g. walking, dance), can use a wide range of different temporal intervals between an arbitrary number of keyframes, and can provide a variety of outputs.  

\textbf{Deep Learning for Motion Synthesis.} Recently, researchers have investigated the use of deep learning approaches for synthesizing plausible but not necessarily physically correct motions. Holden~et~al.~\shortcite{holden2016deep} learn a motion manifold from a large motion capture dataset, and then learn a feedforward network that maps user goals such as a motion path to the motion manifold. 
Li~et~al.~\shortcite{li2017auto} introduced the first real-time method for synthesizing indefinite complex human motions using an auto-conditioned recurrent neural network, but only high-level motion styles can be provided as input.
Holden~et~al.~\shortcite{holden2017phase} later proposed a synthesis approach for character motions using a neural network structure that computes weights as cyclic functions of user inputs and the phase within the motion.
Zhang~et~al.~\shortcite{zhang2018mode} synthesize quadruped motion by dynamically blending groups of weights based on the state of the character. Aristidou~et~al.~\shortcite{aristidou2018deep} break motion sequences into motion words and then cluster these to find descriptive representations of the motion that are called motion signatures and motifs. In contrast, we focus here on synthesizing plausible long-term motion transitions.
Followed by the work of Cai~et~al. ~\shortcite{cai2018deep} which focuses on video completion, recently Yan~et~al.~\shortcite{yanconvolutional} proposed a convolution network that can transform random latent vectors from Gaussian process to human motion sequences, and demonstrated the ability to complete motion between disjoint short motion sequences. But their framework only generates and matches local motion without global motion, and does not apply to the cases of sparsely given individual keyframes. Moreover, they do not provide users explicit control on the style of output motion or consider the body flexibility constraints. We do not compare with their method because it does not provide open-sourced code. Zhang and van de Panne~\shortcite{zhang2018data} use a neural autoregressive model to interpolate between keyframes while following the style of the exemplar corpus. This is demonstrated for a lamp object, unlike our method, which focuses on human motion.  Wang~et~al.~\shortcite{wang2019spatio} focus on learning a deep motion manifold that is suitable for motion synthesis with a long time horizon and motion denoising, which is able to represent the modalities and variance of the motion. 





\textbf{Joint Angle Limits and Forward Kinematics.} Recently, there has been research into the representation of joint angles and the use of forward kinematics within networks. Akhter and Black~\shortcite{akhter2015pose} learn a pose-dependent model of joint angle limits, which generalizes well but avoids impossible poses. Jiang and Liu~\shortcite{jiang2018data} use a physics simulation to accurately simulate human joint limits.  We are inspired by these works and represent local joint angles in a limited range in our network.  Villegas~et~al.~\shortcite{villegas2018neural} use a recurrent neural network with a forward kinematics layer to perform unsupervised motion retargeting. We similarly use a forward kinematics layer in our network. Pavllo~et~al.~\shortcite{pavllo2019} use a quaternionic representation for joint angles and a forward kinematics layer. Zhou~et~al.~\shortcite{zhou2019} showed recently that when the complete set of 3D rotations need to be represented, a 6D rotation representation often performs better: we therefore adopt that angle representation for the root node.




%

%% file: method.tex
\section{Method}

\subsection{Definitions}
In this paper, we use \emph{keyframes} to refer to user-specified root position and local pose at each joint at the specific time. We use \emph{representative frames} to refer to algorithmically extracted representative (or important) poses within a motion sequence. We use \emph{motion DNA} 
to refer to a random sampling of representative frames from the same motion class, embedded to a 1D hidden vector space. This \emph{motion DNA} helps characterize important properties of a motion: conditioning our network on the motion DNA allows the user to create variation in the model's output. This concept is related to the ``motion signatures" of Aristidou~et~al.~\shortcite{aristidou2018deep}. Note that our motion DNA is only roughly analogous to DNA in biology: it helps control the appearance of the overall motion but contains much less information and it is a distinct concept from the keyframes: keyframes precisely control pose at specified times. We use the following notations and a frame rate of 60 fps throughout the paper: 

\emph{Root} refers to the hip joint, which is also the root node. 

$\mathbf{S}=\{\mathbf{S}_t\}_{t=1}^N$: a sequence of synthesized skeleton poses including the global position of the root and the local position of the other joints relative to the root.

$\mathbf{S}_{t_1:t_2}$: a sub sequence of skeleton poses from frame $t_1$ to $t_2$.

$\mathbf{\hat{T}}=\{\mathbf{\hat{T}}_t\}_{t=1}^{\hat{N}}$: a set of representative frames that capture the typical characteristics of a motion class. We rotate the representative frames so that their root joints are all facing the same direction. 

$N$: total number of frames.

$N'$: total number of user's keyframes.

$\phi = \{\phi_t\}_{t=1}^k$: the set of keyframe indices.

$\hat{N}$: total number of the representative frames in Motion DNA

$M$: total number of joints in the skeleton.

$\mathbf{S'}$: a joint position tensor of dimension $N \times 3M$ that contains user-specified keyframes.

$\mathbf{T}^{\mathrm{root}}_{t}$: 3D translation of the root in world coordinates at frame $t$.

$\mathbf{T}^{\mathrm{joints}}_{t}$: 3D translations of the joints relative to the root node in relative world coordinates at frame $t$. The dimension is $M \times 3$.

$\mathbf{R}^{\mathrm{root}}_{t}$: 3D rotation matrix of the root node in world coordinates.

$\mathbf{R}^{\mathrm{joints}}_{t}$: 3D rotation matrices of the joints relative to the parent joints at frame $t$. The dimension is $M \times 3 \times 3$.


\begin{figure}
    \includegraphics[width=3.5in,trim={0.5in 7in 5in 0.5in},clip]{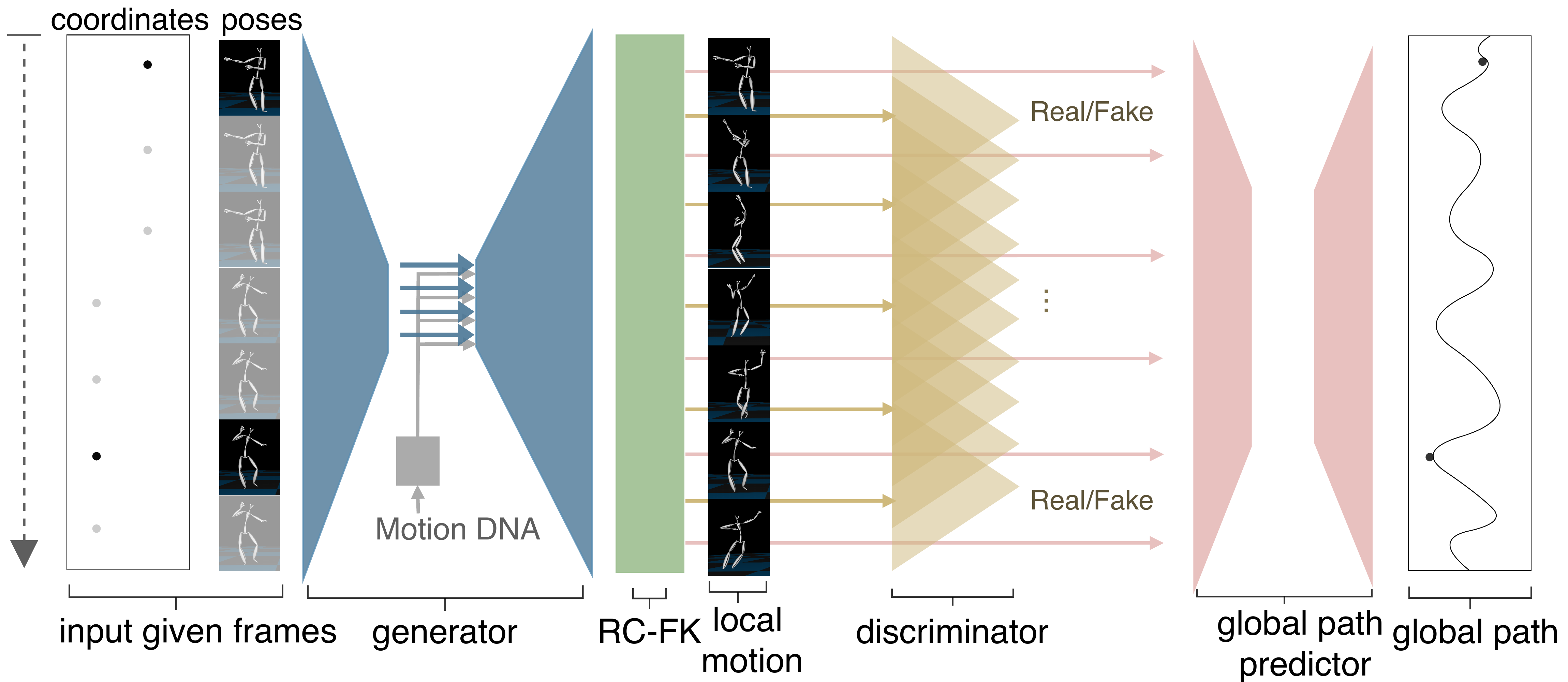}
    \caption{Method Overview. Given user-specified keyframes and the corresponding mask, we generate the local motion of every joint and every frame in a rotation representation. This is done by a fully convolutional encoder and decoder, shown in blue, which obtain extra information about style and appearance from the Motion DNA component. We then use Range-Constrainted Forward Kinematics (RC-FK) module to obtain the local joint positions. During training, a fully convolutional discriminator (shown in yellow) is used to assess realism of the produced subsequences. Finally, the local motion is passed to a global path predictor to estimate the global path of the root joint. A loss is applied to match the pose closely at the keyframes, which are indicated by black dots.}
    \label{fig:overall_method}
\end{figure}
\subsection{Method Overview}
Figure~\ref{fig:overall_method} shows our system design. 
At a high level, we use a 1-D CNN to first predict the local pose of all frames conditioning upon user-provided sparse keyframes, and then from the local pose we use another 1-D CNN to predict the global pose. We first explain the motivation of such a two-stage approach.

%
The problem of long-term inbetweening can be formulated as:
\begin{equation}
\begin{aligned}
\quad & \mathbf{S} = \mathbf{G}_0(\mathbf{S}^{'}_{\phi})\\
\quad & \min_{\mathbf{G}_0}{ \sum_{t=1}^{N-n} {\mathcal{L}_\mathrm{realism}(\mathbf{S}_{t:t+n})}}\\
\quad & \mathrm{s.t.} \forall t\in \phi, \mathbf{S}_{t} = \mathbf{S}^{'}_{t}    \\
\end{aligned}
\end{equation}
where $\mathbf{S}$ is the synthesized motion sequence that contains all joint information, $\mathbf{S}^{'}$ is the user-specified pose, which is defined for the keyframes at times $\phi$, $\mathbf{G}_0$ is a generative model, and $\mathcal{L}_\mathrm{realism}$ is some loss that encourages realism. For simplicity, we use sub-sequences containing $n$ frames to measure realism. 
Note that given very sparse keyframes, there might be multiple good solutions. 

One approach to measure realism is to use a generative adversarial network (GAN). A GAN could be conditioned on the user-specified keyframes $\mathbf{S}^{'}$, uses a discriminator to measure the realism, and uses a generator to hallucinate complex motion while simultaneously matching the keyframes and deceiving the discriminator. 
However, we found that empirically, training a GAN to generate global and local motions together is much more unstable than training it to generate local motions only. This is partially because the output realism is very sensitive to global motion, e.g. 10 degree tilt on root while walking leads to impossible motion, but the local poses are still good. Predicting local motions alone significantly simplifies the problem and allows the network to focus on generating realistic local poses.
Fortunately, we found that the global motion can be inferred from the synthesized local motions using a pre-trained and fixed network that we call the ``global path predictor". 
%
%
Thus, we design a two-stage method to first generate the local motions and then the global path. We reformulate the optimization as:
\begin{equation}
\begin{aligned}
\quad & \mathbf{T}^{\mathrm{joints}} = \mathbf{G}(\mathbf{S}^{'}_{\phi}),  \mathbf{T}^{\mathrm{root}} = \mathbf{\Upsilon}(\mathbf{T}^{\mathrm{joints}})\\
\quad & \mathbf{S}_{t} = \{\mathbf{T}^{\mathrm{root}}_{t}, \mathbf{T}^{\mathrm{joints}}_{t}\}\\
\quad & \min_{\mathbf{\Upsilon}, \mathbf{G}} \sum_{t=1}^{N-n} {\mathcal{L}_\mathrm{realism}\left( \mathbf{T}^{\mathrm{joints}}_{t:t+n}\right)} + \gamma \sum_{t\in \phi} {\|\mathbf{S}_{t} - \mathbf{S}^{'}_{t}\|^2}\\
\end{aligned}
\end{equation}

The first network $\mathbf{G}$ is a generator network that is conditioned on keyframes and outputs the local motion $\mathbf{T}^{\mathrm{joints}}_t$. The second network $\mathbf{Y}$ is a global path predictor, which is conditioned on the local motion and predicts the global motion, that is the velocities $\mathbf{dT}^{root}_t = \mathbf{T}^{root}_{t+1} - \mathbf{T}^{root}_t$, which can be summed to recover the global motion $\mathbf{T}^{root}_t$. Because the global path predictor is conditioned on the local motion, and the local motion is conditioned on the keyframes, this means the global path predictor indirectly receives information related to the keyframes. We can also view the split into local and global motion as being related to learning disentangled representations (e.g. ~\cite{gonzalez2018image}): by forcing one network to generate only local pose and the other network to integrate up these poses into global motion, we simplify the problem into two coupled, but more specialized learning tasks. For $\mathcal{L}_\mathrm{realism}$, we use a GAN loss (\sect{sec:generator_and_discriminator}). Finally, we use an $L^2$ loss to measure the mismatch between the generated frame $\mathbf{S}_{t}$ and the user's keyframe $\mathbf{S}^{'}_{t}$. 
In practice, we first train the global path predictor $Y$ using the motions in the CMU motion dataset~\cite{cmu}. After pretraining the global path predictor, we then freeze its weights and train the generator $\mathbf{G}$. 

Most deep learning based motion synthesis works~\cite{Harvey:2018:Recurrent, MordatchLAPT15, villegas2018neural} use recurrent neural network (RNN) which is specifically designed for modeling the dynamic temporal behavior of sequential data such as motion. However, traditional RNN only uses the past frames to predict future frames. 
Harvey and Pal~\shortcite{Harvey:2018:Recurrent} proposed Recurrent Transition Network (RTN), a modified version of LSTM, which generates fixed-length transitions based on a few frames in the past and a target frame in the future. However, RTN cannot process multiple future keyframes spaced out in time and is therefore only suitable for modeling simple locomotion which does not need to take into account a global context specified by multiple keyframes. 
In contrast, a 1-D convolutional network structure that performs convolutions along the time axis is more suitable for our scenario, because multiple keyframes can influence the calculation of a transition frame given large enough network receptive field. 
Thus, we use 1-D convolutional neural network for our global path predictor, motion generator and discriminator. 

As illustrated in Figure~\ref{fig:overall_method}, given the global coordinates of the root and the local positions of other joints at keyframes (black dots and frames in the leftmost region), a generator synthesizes the local pose in rotation representation for all the frames. A motion DNA vector is additionally fed into the generator to enable varied outputs and encourage the output to match in style. Next, a Range-Constrained Forward Kinematics (RC-FK) layer translates the local pose in rotation representation to 3D positions, which are then fed into the global path predictor to generate the trajectory of the root. We developed the RC-FK layer to remove discontinuities in the angle representation, which can harm learning~\cite{zhou2019}. The remaining sections will discuss each of the components in details.

\subsection{Range-Constrained Motion Representation}
\label{sec:RC_FK}
As a reminder, throughout our network, following Zhou~et~al.~\shortcite{zhou2019}, we would like to use continuous rotation representations in order to make the learning process converge better. A skeleton's local pose at a single frame can either be defined by the local joint translations $\mathbf{T}^{\mathrm{joints}} = \{\mathbf{T}_i\}_{i=1}^M$ or hierarchically by the relative rotation of each joint from the parent joint $\mathbf{R}^{\mathrm{joints}} = \{\mathbf{R}_i\}_{i=1}^M$. By denoting the initial pose as $\mathbf{T}^{\mathrm{joints}}_{0}$ (usually a T pose) and the parent index vector $\varphi_{\mathrm{child}\_\mathrm{id}}=\mathrm{parent}\_\mathrm{id}$, we can map the rotation representation to the translation representation by a differentiable Forward Kinematics function $FK(\mathbf{R}^{\mathrm{joints}}, \mathbf{T}^{\mathrm{joints}}_{0}, \varphi)= \mathbf{T}^{\mathrm{joints}}$ ~\cite{villegas2018neural}.

%
Many public motion datasets do not capture the joint rotations directly, but instead use markers mounted on human performers to capture the joint coordinates, which are then transformed to rotations. Due to insufficient number of markers, the imprecision of the capture system, and data noise, it is not always guaranteed that the calculated rotations are continuous in time or within the range of joint flexibility. 
In contrast, joint positional representation is guaranteed to be continuous and thus are suitable for deep neural network training~\cite{zhou2019}. Therefore, we use the joint translations as the input to our generator and discriminator. However, for the motion synthesized by the generator network, we prefer to use a rotation representation, since we want to guarantee the invariance of the bone length and restrict the joint rotation ranges according to the flexibility of the human body. These kinds of hard constraints would be more difficult to enforce for a joint positional representation.


We now explain the rotation representation we use for the local joints. We define a different local Euler angle coordinate system for each joint with the origin at the joint, x and y axis perpendicular to the bone and the z axis aligned with the child bone (and if a joint has multiple child bones, then the mean of the child bones). 
Due to joint restrictions on range of motion, the rotation angles around each axis should fall in a range, $[\alpha, \beta]$, where $\alpha$ is the minimal angle and $\beta$ is the maximal angle. We have the neural network output the Euler angles $v=(v_x, v_y, v_z)^T\in \mathbb{R}^3$ for each joint except for the root joint and then map them to the feasible range by: 
\begin{equation}
u = \left[\begin{array}{c}
u_x \\
u_y \\
u_z \\
\end{array}
\right]
=
\left[\begin{array}{c}
(\alpha_x-\beta_x)\tanh(v_x)+(\alpha_x-\beta_x)/2 \\
(\alpha_y-\beta_y)\tanh(v_y)+(\alpha_y-\beta_y)/2 \\
(\alpha_z-\beta_z)\tanh(v_z)+(\alpha_z-\beta_z)/2 \\
\end{array}
\right]
\end{equation}
We can move from the relative rotation of the joints to world coordinates in the usual way for forward kinematics by composing the relevant 4x4 homogeneous rotations and translation matrices.


Although it is convenient to enforce the rotation range constraint, the Euler angle representation can have a discontinuity problem, which can make it hard for neural networks to learn~\cite{zhou2019}. 
The discontinuity happens when the second rotation angle reaches $(-90\degree$ or $ 90\degree)$, which is known as the Gimbal lock problem. 
Fortunately, each body joint except the hip joint has at least one rotation axis along which the feasible rotation range is smaller than $(-90\degree, 90\degree)$. Thus, we can avoid Gimbal lock by choosing the order of rotations accordingly. 
Therefore, we compute the rotation matrix of the left and the right forearms in the $Y_1X_2Z_3$ order and the other joints in the $X_1Z_2Y_3$ order as defined by Wikipedia for the intrinsic rotation representation of Tait-Bryan Euler angles~\cite{wiki:euler_angles}. For the hip, since there is no restriction in its rotation range, to avoid discontinuities, we use the continuous 6D rotation representation defined in~\cite{zhou2019}.

The detailed information regarding the range and order of the rotations at each joint can be found in the appendix.


\subsection{Global Path Predictor}
\label{sec:global}
We infer the global translations of the root using a 1D fully convolutional network which takes the synthesized local joint translations as input and outputs the velocity sequence of the root joint $\mathbf{dT}^{root}_t = \mathbf{T}^{root}_{t+1} - \mathbf{T}^{root}_t$. The architecture is in Table \ref{tab:network_architectures} of the appendix. The network is trained with the ground truth motions in CMU Motion Capture dataset~\cite{cmu}. 
Between any two times $t_0$ and $t$, we can compute the displacement in the root position as:
\begin{equation}
    \mathbf{\Delta}^{root}_{t_0->t} = \sum_{i=t_0}^{t}\mathbf{dT}^{root}_i
\end{equation}

To prevent drifting due to error accumulation in the predicted velocity, we penalize root displacement error in each $n$-frame interval, $n=1,2,4,...,128$.
\begin{equation}
    L = \frac{1}{8} \sum_{q=0}^{7} \frac{1}{N-n}\sum_{t=1}^{N-n} ||\mathbf{\Delta}^{root}_{t->t+n} - \mathbf{\Delta}'^{root}_{t->t+n}||^2
\end{equation}
where by abuse of notation, $n=2^q$, $\mathbf{\Delta}^{root}_{t->t+n}$ is the root displacement from frame $t$ to frame $t+n$ calculated using the predicted $\mathbf{dT_{t+1,..., t+n}}$ and $\mathbf{\Delta'}^{root}_{t->t+n}$ is the corresponding ground truth displacement, that is, the difference between the ground truth frames.

\subsection{Local Motion Generation}
\label{sec:generation}
\begin{figure}
    \includegraphics[width=3.3in,trim={0 1cm 0 0},clip]{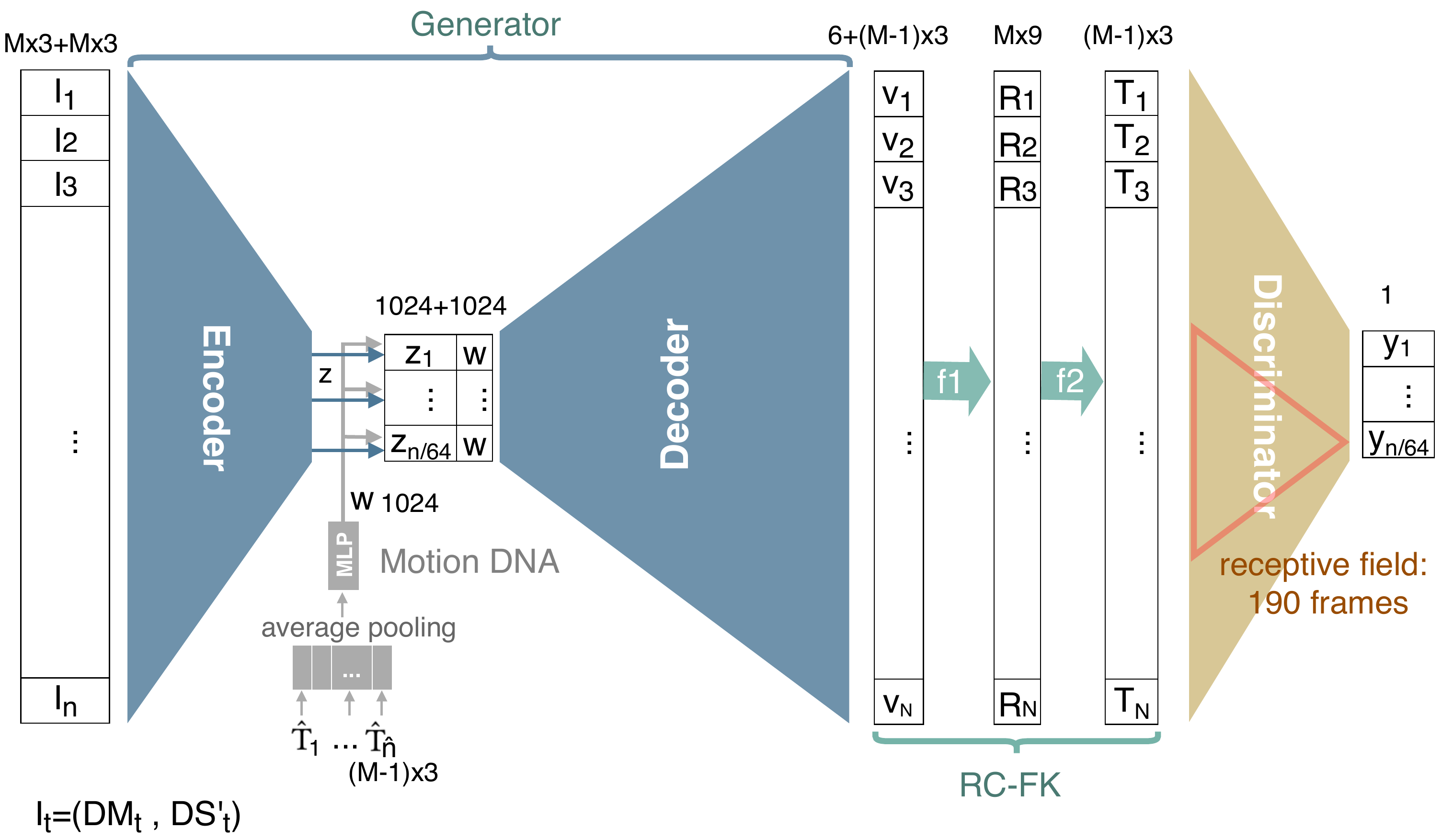}
    \caption{Network Architecture for Local Motion Generation. As explained in Section~\ref{sec:RC_FK}, $\mathbf{V}_t$ contains a 6-D vector that represents the rotation of the root and (M-1) 3-D vectors that represent the rotation of the other joints. $\mathbf{R}_t$ contain the 3x3 rotation matrices for all the joints. $\mathbf{T}_t$ and $\mathbf{\hat{T}}_t$ contain the local translation vectors for all the joints except for the root joint. 
$\mathbf{w}$ is concatenated with each $\mathbf{z}_t$ to be fed into the decoder.}
    \label{fig:local_GAN}
\end{figure}

The local motion generation module takes as input a sequence of vectors that capture the locations and values of the keyframes and outputs the local motion for the whole sequence. 
As shown in Figure \ref{fig:local_GAN}, it is composed of five parts: 
(1) an encoder to encode the input sequence into a sequence of 1-D latent vectors $\mathbf{z}_t \in \mathbb{R}^{1024}$, 
(2) a Motion DNA encoder to encode a set of randomly picked representative frames $\mathbf{\hat{T}}$ into a 1-D latent vector $\mathbf{w}\in \mathbb{R}^{1024}$, 
(3) a decoder to synthesize the local motion of the entire sequence in the rotation representation, 
(4) a Range-Constrained Forward Kinematic (RC-FK) module to transform the local motion from the rotation representation back to the positional representation $\mathbf{S}$, and
(5) a discriminator to encourage realism of the generated sub-sequences. 

In the following subsections, we first explain our dense keyframe input format, then we discuss the architecture of our generator and discriminator, and the losses we use at training time. We then explain how our Motion DNA works.

\begin{figure}
    \includegraphics[width=3.3in]{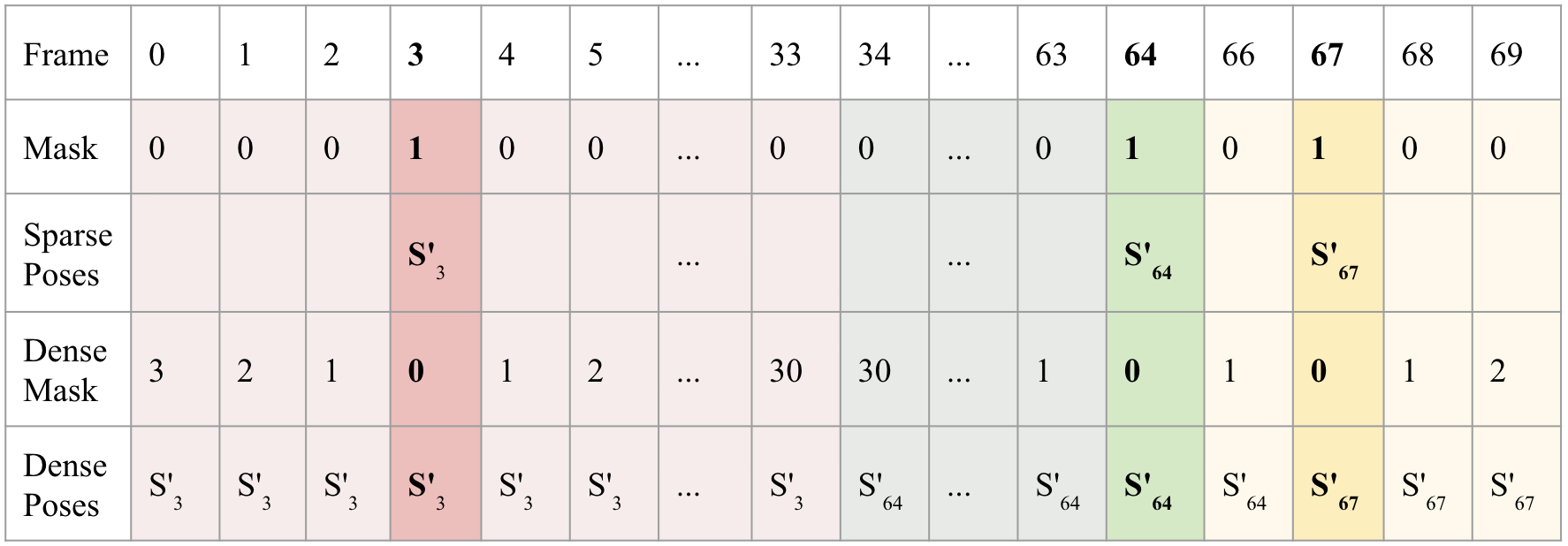}
    \caption{Input Format for Local Motion Generator. The first row contains the frame indices. The second and third rows show the sparse input format, and the fourth and fifth rows show the dense input format. 1s and 0s in the second row indicate the presence and absence of keyframes. $S'_{\phi}, \phi=3, 64, 67$ are the poses at user-specified keyframes.
    }
    \label{fig:input_format}
\end{figure}

\paragraph{Dense Keyframe Input Format.}
\label{sec:input_sequence_format}
The local motion generator receives sparse input keyframes that are specified by the user. We could format these keyframes in either a sparse or dense format: we will explain the sparse format first, because it is easier to explain, but in our experiments, we prefer the dense format, because it accelerates convergence. In the case of the \emph{sparse input format}, the inputs are: (1) a mask of dimension $N \times 3M$ that indicates at each frame and each joint, whether the joint's coordinate values have been specified by the user, and 
(2) a joint position tensor $\mathbf{S'}$ of dimension $N \times 3M$ that contains user-specified world-space joint positions at keyframes $\phi$ and zero for the rest of the frames. These are shown in the second and third rows in Figure~\ref{fig:input_format}. This sparse input format can encode the pose and timing information of the users' keyframes and is flexible for cases of partial pose input, e.g. the user can give the root translation without specifying the local poses. However, since the keyframes are sparse, the input sequence usually contains mostly zeros, which results in useless computation in the encoder. Thus, we prefer the \emph{dense input format}, which is shown in the fourth and fifth rows of Figure~\ref{fig:input_format}. In the dense input format, each frame in $\mathbf{S'}$ gets a copy of the pose from its closest keyframe. 
The mask is also replaced with a distance transform, so the values at the keyframes are zero, and the values at the other frames reflect the absolute difference in index between that frame and a closest keyframe. 
The dense input format increases the effective receptive field of the generator, since with keyframes duplicated at multiple input locations, each neuron in the bottleneck layer can see a larger number of keyframes. Empirically, this representation also leads to a faster network convergence. 


\paragraph{Generator and Discriminator Architecture.}
\label{sec:generator_and_discriminator}
The encoder contains six convolutional layers that reduce the input dimension by $2^6$ = 64 times, and the decoder contains eight convolutional layers and six transpose convolutional layers to upsample the latent representation to the original input dimension. The size of the receptive field at the bottleneck layer is 318 (5.3 seconds) frames. With the dense input format introduced in Section~\ref{sec:input_sequence_format}, the effective receptive field becomes 636 frames (10.6 seconds): this is theoretically the longest  time interval between any neighbouring keyframes, because otherwise the synthesized transition at a particular frame will not be dependent on the two adjacent keyframes. 

The discriminator takes as input the entire synthesized sequence, passes it to seven convolutonal layers (6 layers perform downsampling and 1 layer maintains the input dimension) and outputs $N/64$ scalars $y_t$. The discriminator looks at all the overlapping sub-sequences ($N/64$ sub-sequences of length 64 frames each) and predicts whether each of them looks natural (one output scalar per sub-sequence). The receptive field of the discriminator is 190 frames: this is the length of the sub-sequence that each $y_t$ is responsible for. 

During experiments, we found that generators using vanilla convolutional layers cannot precisely reconstruct the input keyframes and tend to to have mode collapse, by synthesizing similar motions for different inputs. We hypothesize that this is because the network is too deep and the bottleneck is too narrow. To this end, we introduce a modified residual convolution layer and a batch regularization scheme:

\begin{equation}
\begin{aligned}
\quad & \mathbf{x}'=\text{conv}(\mathbf{x})\\
\quad & \text{residual}(\mathbf{x}) =\text{PReLU}(\text{affine}(\mathbf{x}')) + \sqrt{1-\sigma}\mathbf{x} \\
\quad & \mathbf{y}=\sqrt{\sigma}\text{residual}(\mathbf{x}) + \sqrt{1-\sigma}\mathbf{x} \\
\end{aligned}
\label{eqn:our_residual_convolution}
\end{equation}

Here $\mathbf{x}$ is the input tensor and $\mathbf{y}$ is the output. PReLU() is an activation function called Parametric Rectified Linear Unit~\cite{He:2015}. affine() is a learned affine transformation, which multiplies each feature elementwise by a learned scale-factor and adds to each feature elementwise a learned per-feature bias. conv() stands for the convolutional filter with learnable kernels. The residual ratio $\sigma$ is a scalar from 0 to 1. 
In our network design as shown in the appendix, we set $\sigma$ in a decreasing manner. The intuition here is that the layers closer to the output mainly add low-level details, and therefore we want to pass more information directly from the previous layer, especially near keyframes. 

\paragraph{Losses for GAN and Regularization.}

We applied the Least Square GAN Losses~\cite{Mao2017LeastSG} on our generator and discriminator. 
We use 1 as the real label and -1 as the fake label and therefore $y_t^{real}$ should be close to 1 and $y_t^{fake}$ should be close to -1. 
\begin{equation}
\begin{aligned}
\quad & L^D = 1/(N/64) (\sum_{t=1}^{N/64}{(y_t^{real}-1)^2} + \sum_{t=1}^{N/64}{(y_t^{fake}+1)^2})
\end{aligned}
\end{equation}

For the generator, we want the corresponding $y_t$ to be close to neither -1 nor 1. According to the Least Square GAN paper~\cite{Mao2017LeastSG}, one can use any value in [0, 1), and based on experiments, we chose 0.2361, so we have:
\begin{equation}
\begin{aligned}
L^G_{adv} = 1/(N/64)\sum_{t=1}^{N/64}{(y_t^{fake}-0.2361)^2} 
\end{aligned}
\end{equation}

Before we feed synthesized local motion sequences $\mathbf{S}$ into the discriminator, we replace the pose at the keyframe locations with the input pose, because we found that this encourages the network to generate smooth movement around the keyframes. 

For our problem, we found that standard batch normalization~\cite{ioffe2015batch} does not work as well as a custom batch regularization loss. This appears to be because when using batch normalization, different mini-batches can be normalized differently due to different mean and standard deviations, and there may also be some difference between these values between training and test times. We apply the following batch regularization loss to each modified residual convolution layer in the generator and discriminator:
\begin{equation}
    L_{\mathrm{br}} = 1/D \sum_{\mathbf{x}\in\{\mathbf{X}'\}} (||\mathrm{E}(\mathbf{x}')||^2_2 + ||\log(\sigma(\mathbf{x}'))||^2_2)
\end{equation}
Here $\{\mathbf{X}'\}$ is the set of outputs of the convolution layers in the residual block, D is the total dimension of $\{\mathbf{X}'\}$, $\mathrm{E}$ and $\sigma$ refer to the sample mean and standard deviation of the features. We also found empirically that this loss helps encourage the diversity of the output motions and avoid mode collapse.

\paragraph{Losses for Keyframe Alignment.}
%
To further enforce the keyframe constraints, we use L2 reconstruction loss on the position vectors of all non-root joints at keyframes. 

\begin{figure}
 \includegraphics[width=0.7\linewidth]{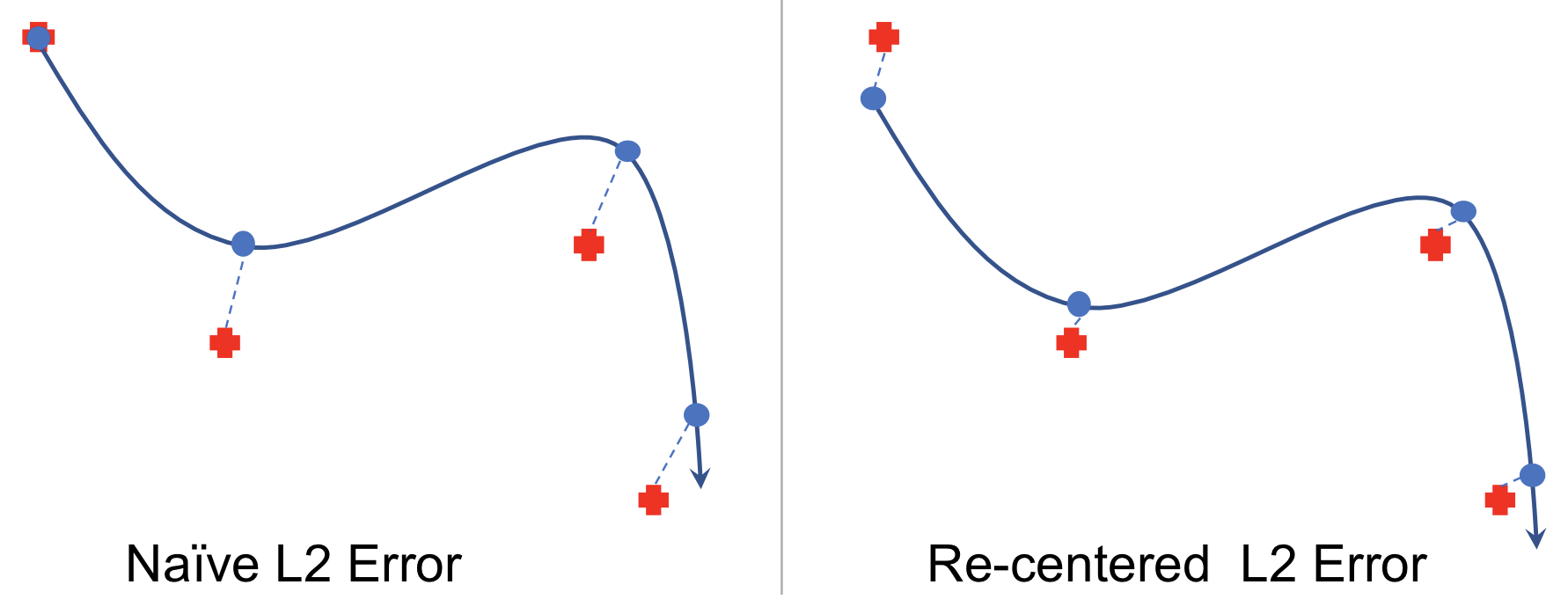}
    \caption{Root translation error calculation. The blue path is the synthesized root path, and blue dots are the synthetic root positions at the keyframes. Red crosses indicate the input root positions at the keyframes. 
}
    \label{fig:root_distance}
\end{figure}

However, for the root (hip) node, applying such a naive L2 distance calculation is problematic. We show one example of why this is the case in Figure~\ref{fig:root_distance}. In that example, the output root trajectory matches well the input in terms of direction, but a large error is reported because the synthesized and target point clouds are not well-aligned, due to a cumulative drift in prediction over time. 
To fix this, we recenter the output path by placing the mean of the output root trajectory $\mathbf{T^c}$ at the mean of the input root trajectory $\mathbf{T'^c}$, where both means are taken only over keyframes. This is shown in the right part of Figure~\ref{fig:root_distance}. The updated loss function becomes:
\begin{equation}
    L_{root} = \frac{1}{N'}\sum_{t\in\{\phi\}}||(\mathbf{T}^{root}_{t}-\mathbf{T^c})
    -(\mathbf{T'}^{root}_{t} - \mathbf{T'^c})||^2_2
\end{equation}

This is equivalent to performing a least-squares fit over all translations. Least-squares fits over all rigid transforms have also been studied~\cite{arun1987least,lorusso1995comparison}. We opted to not fit the rotational components because such a rotation would induce changes in the orientation for the local motion, which could make the local motion harder to learn, since the target orientation would not match the input keyframes and would depend upon this downstream normalization. 

\subsubsection{Motion DNA}
To encourage the network to hallucinate different motion sequences following the same keyframe constraints, we introduce a novel conditioning method that we call Motion DNA. By changing the Motion DNA, users can provide some hints about the type of motion to be generated. 

We were motivated to create such a scheme because we noticed that motion within the same class can sometimes exhibit dramatically different styles and it is impossible to classify all types of motions by limited labels.  For example, in the CMU Motion Capture dataset, a class called ``animal behaviors" contains human imitations of different animals such as elephants and monkeys, but their motions don't share any common features.


Instead of using labels, we found that a small set of representative frames of a motion sequence contain discriminative and consistent features that can well represent the characteristics of a particular type of motion. For example, the yellow skeletons in the fifth line in Figure \ref{fig:DNA_effect} are the representative frames extracted from a motion sequence, and it is not hard to tell that they are extracted from a martial arts sequence. We identify representative frames as those that have the maximum joint rotation changes from their adjacent frames. The extracted representative frames can be injected into the generator to control the characteristics of the generated motion. The advantage of using the representative frames as the controlling seeds is that it can be applied to general types of motions and the user can explicitly define their own combination of the representative frames.

To encode the set of representative frames into a Motion DNA, we use a shared multilayer perceptron network to process each representative frame $\mathbf{\hat{T}}_j, j=0,1,...,\hat{N}$ and then apply average-pooling and an additional linear layer to obtain the final feature vector $w$. We concatenate $w$ at the end of each $z_t$ and use that as input to the decoder.   

To encourage the local motion generator to adopt the Motion DNA to generate variations without conflicting with the keyframe constraints, we design two losses to encourage the generator to place representative poses from the Motion DNA into the synthesized motion at suitable locations. 

The first loss $L^G_{DNA_1}$ encourages all of the representative poses to show up in the output sequence. For each representative frame $\mathbf{\hat{T}}_j$, we find the most similar frame in the synthesized motion $\mathbf{S}$ and compute the error $e_j$ between the pair. 
\begin{equation}
\begin{aligned}
\quad & L^G_{DNA_1} = 1/\hat{N} \sum_{j=0,..,\hat{N}} e_j
\quad & e_j = \min_{\mathbf{S}_i \in \mathbf{S}} ||\Lambda(\mathbf{S}_i) - \mathbf{\hat{T}}_j||^2_2
\end{aligned}
\label{eqt:DNA1}
\end{equation}
$\Lambda(\mathbf{S}_i)$ is the local pose of $\mathbf{S}_i$ without root rotation. Likewise, the representative poses $\mathbf{\hat{T}}_j$ also does not contain root rotations. In this way, we only care about the relative posture and not direction.

Because $L^G_{DNA_1}$ does not consider how the representative frames are distributed among the generated frames, it is possible that the representative poses all placed in a small time interval, which is not desirable. 
Therefore, we use a second loss $L^G_{DNA_2}$ to encourage the representative poses to be distributed  evenly throughout the entire synthesized sequence. 
Specifically, we cut the whole synthesized sequence at the keyframe points, and divide it into $K$ intervals, $u_k, k=0,1,...,K-1$, with each interval having the length of $N_k$, so $S = [S_{u_0}, S_{u_1}, ..., S_{u_{K-1}}]$. If $N_k$ is shorter than 3 seconds, we consider it is hard to place representative poses in it, otherwise we want the synthesized sub-sequence to contain frames that closely resemble $\hat{N}_k = \lfloor{N_k / 3}\rfloor$ input representative frames. In practice, for each frame $\mathbf{S}_{k,i}$ in $u_k$, we look for one of the representative poses in $\mathbf{\hat{T}}$ that is closest to it, and compute their error as $e'_{k,i}$. We consider the $\hat{N}_k$ frames with the lowest errors to be the frames where the representative poses should show up, and collect their errors. $L^G_{DNA_2}$ is defined as the average of those errors: 
\begin{equation}
\begin{aligned}
\quad & L^G_{DNA_2} = \frac{1}{\sum{\hat{N}_k}} \sum_{e'_{k,i}\in E'_k} e'_{k,i}
\\
\quad & e'_{k,i} = \min_{\mathbf{\hat{T}_j} \in \mathbf{\hat{T}}} ||\Lambda(\mathbf{S_{k,i}}) - \mathbf{\hat{T}}_j||^2_2
\end{aligned}
\label{eqt:DNA2}
\end{equation}
,$e'_{k,i}$ is the minimum error between the $i$-th local pose in the interval $u_k$ and all the representative frames. $E'_k$ is the set of the lowest $\hat{N}_k$ errors in the interval $u_k$.





\subsection{Post-processing}
\label{sec:post}
Though we use L2 reconstruction loss to enforce the keyframe constraints, the output sequence might not exactly match the input at keyframe locations. When needed, simple post processing can be applied to achieve exact keyframe matching~\cite{Harvey:2018:Recurrent}. 
With small keyframe mismatches in our synthesized motion, the adjustments are nearly undetectable.

\subsection{Dataset and Training}
\label{sec:train_details}

We construct our training dataset based on the CMU Motion Capture Dataset \cite{cmu}. We only use motion clips in which the entire motion is performed on a flat plane. We automatically filtered out noisy frames (see appendix). The final dataset contains 79 classes of motion clips of length ranging from 4 seconds to 96 seconds. We use 1220 clips for training and 242 clips for testing.

For training, we first train the Global Path Predictor model (\sect{sec:global}), fix its weights and then train the Local Motion Generator model (\sect{sec:generation}). 
For each generator update, we update the discriminator once. 
We feed fixed-length 2048-frame sequences to the generator with a batch size of 4. We train the discriminator using 16 real sequences of length ranging from 256 to 1024 frames in each batch. We use the RMSprop optimizer with the default parameters from Pytorch~\cite{paszke2017automatic}. Other details about training parameters and frame sampling can be found in the Appendix.








%% file: experiments_new.tex
\section{Experiments and Evaluation}
We conducted various experiments to evaluate the different components of our system. We first evaluated the effectiveness of our Range-Constrained motion representation. We then measured the accuracy of the global path predictor by comparing its results to the ground truth. We also evaluated the local motion generator in two ways: (1) quantitatively, we measured how well the generated motion follows the input keyframe constraints, and (2) we conducted user studies to evaluate the quality of the synthesized motion given different keyframe placement strategies: we compare our approach with a prior work~\cite{Harvey:2018:Recurrent}. Finally, we examined the effectiveness of using Motion DNA to generate motion variations.

\subsection{Effect of Range-Constrained Motion Representation}
We verify the effect of our Range-Constrained motion representation by training the same Inverse Kinematic networks as in \cite{zhou2019}. We show in Table \ref{tab:RP} that our representation and RC-FK layer outperform the other rotation representations. 
\input{table/eval_representation.tex} 

\begin{figure}
\vspace{-0.8cm}
    \includegraphics[width=0.85\linewidth]{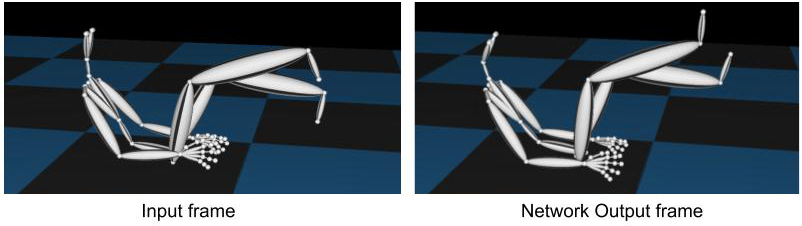}
    \vspace{-0.3cm}
    \caption{Input frame and output frame from the network with a RC\_FK layer.}
    \label{fig:RC_FK_output}
\end{figure}

Additionally, in practice, both the ground truth training data and user's given keyframes can have bad poses, and a trained network with the RC-FK layer has the ability to correct those artifacts. For example, in Figure~\ref{fig:RC_FK_output}, left is an input frame picked from the real motion dataset where the feet are bent the wrong way. Right is our network output where the feet artifacts are fixed.

\subsection{Accuracy of the Global Path Predictor}
We trained the Global Path Predictor model until it reached the precision as shown in Table \ref{tab:hip_prediction_error}. The path prediction error is 0.7 cm per frame, and only 5.5 cm after 2 seconds (128 frames) of motion. This shows that our global path prediction is quite accurate for motion prediction over a few seconds.
\input{table/eval_path_prediction.tex}

\subsection{Keyframe Alignment}
\label{sec:exp_keyframe_alignment}
We next evaluate how well our network is able to match the keyframe constraints. We first pretrain and fix the Global Path Predictor Network. We then use this to train the in-betweening network for 560,000 iterations and evaluate how well the synthesized motion aligns to the input keyframes. The input keyframes are given as the local joint coordinates plus the global root positions at some random frames. For testing, we generate 200 sets of 2048-frame-long motion sequences using random keyframes and DNA poses with the sampling scheme described in Section \ref{sec:train_details}. We calculate the mean L2 distance error of the global root joints and the local joints at the keyframes. As shown in the last two rows of Table \ref{tab:user_study_results}, the alignment error is sufficiently small that it can be easily removed by the post-processing in a way that is hardly perceptible.

We show an example rendering of the synthesized frames overlaid with the input keyframes in Figure~\ref{fig:DNA_effect}. The faint pink skeletons are the input keyframes, while the blue skeletons are synthesized by our method. It can be seen that the synthesized frames smoothly transition between the keyframes, which are specified at frame 4 and 11, and faithfully match the keyframes. More results for longer term inbetweening tasks can be found in the supplementary video.

\begin{figure}
    \includegraphics[width=3.2in]{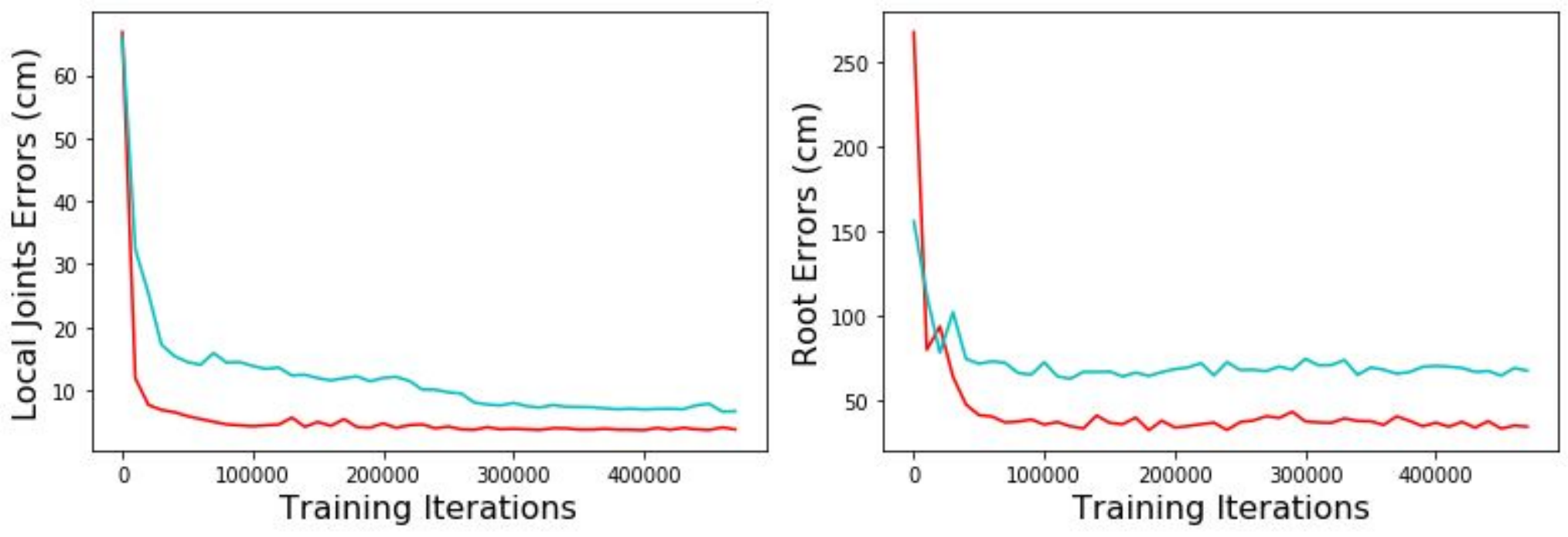}
    \vspace{-0.2cm}
    \caption{Mean L2 errors (cm) of local joints (left) and root joint (right) at keyframes throughout the training process. Blue lines refer to the result of using the sparse input format. Red lines refer to the dense input format.}
    \label{fig:eval_sparseORdense}
\end{figure}

\textbf{Dense vs. Sparse Input. }
In Section \ref{sec:input_sequence_format}, we mentioned that using a dense input format can accelerate the training. We show this effect in Figure \ref{fig:eval_sparseORdense} by evaluating the root and local joint alignment errors at the input keyframes throughout the training iterations. For each test, we generate 2048-frame-long motion sequences from 50 sets of randomly sampled keyframes and Motion DNAs, and we calculate the mean L2 errors between the input and output poses. The results show that the dense input format not only converges to lower errors but also converges much faster than the sparse input format. 

\subsection{Comparison of Different GAN Architectures}
\label{sect:comparison_gan}

\begin{figure}
    \vspace{-0.5cm}
    \includegraphics[width=\linewidth]{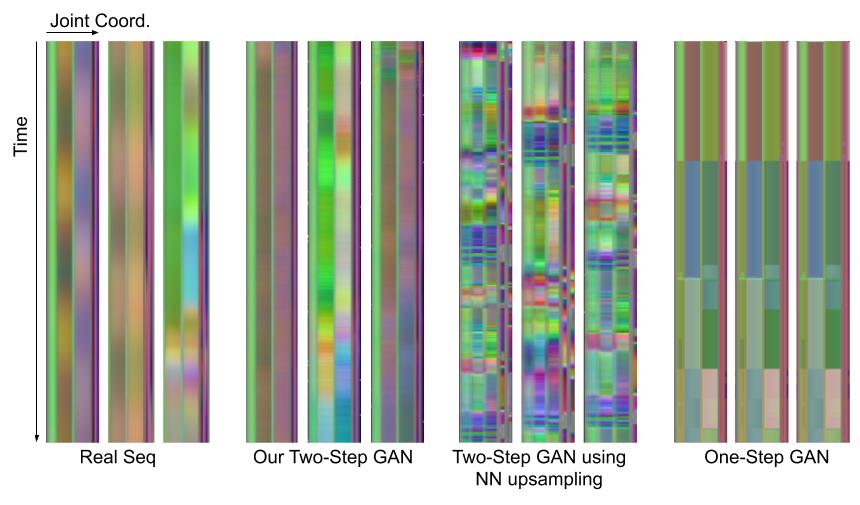}
    \vspace{-0.8cm}
    \caption{Visualized synthetic and real motion data. Please see \sect{sect:comparison_gan} for an explanation of the visualization. Since all joints and a range of times are visualized in the subfigures, it is difficult to imagine the motion from this visualization, but the key takeaway is that the frequency characteristics of each joint are completely wrong for the alternative GAN approaches.}
    \label{fig:GAN_architecture_comparison}
\end{figure}

We compare the result of using different GAN architectures trained by 65,000 iterations. In Figure~\ref{fig:GAN_architecture_comparison}, each block contains three randomly picked motion sequences visualized by the coordinate values of all joints through time. From the first and second block, we can see that the real motion and our two-step GAN model's output motion have similar smooth patterns. The third block is the result of replacing the transpose convolution layers in our model with nearest neighbor upsampling layers. Although the using of upsampling layers were more popular for GAN models in the 2D image generation domain, empirically we found it leads to very noisy output and has the random mosaic patterns. The fourth block is the result of using a one-step GAN that infers both local motion and global motion simultaneously. This network diverges to bad and repeated outputs.

\subsection{Runtime Performance}
We conduct all our experiments on a machine with an Intel{\tiny{\textregistered}} Core$^TM$ i9-7900X CPU with 3.30GHz, 94GB memory, and a Titan Xp GPU. Our code is implemented on an Ubuntu 16.04 and uses Python 3.6 and Torch 0.4. We tested the computation time for generating motion sequences at different lengths. Table \ref{tab:speed} gives the average computation time for the network inference (forward pass) and post-processing (to enforce keyframes matching and smooth output motion), and shows that our method can achieve real-time performance for generating 15 seconds of motion. For longer motions, the system is still fast, using only within 0.4 second for generating 120 seconds of motion.

\input{table/eval_speed.tex}

\subsection{Motion Quality}
\label{sec:motion_quality}
We evaluated the quality of the synthesized motions from our model with a user study. We used the trained model to generate 50 34-second-long (2048 frame) motion sequences using the keyframe and DNA sampling scheme described in Section \ref{sec:train_details}. Next, we randomly and uniformly picked 100 4-second-long subsequences from these sequences. We ran an Amazon Mechanical Turk~\cite{AMT} study. We showed human workers pairs of 4-second real motions and 4-second synthetic motions, with randomization for which motion is on the left vs right. We asked the user to choose which motion looks more realistic, regardless of the content and complexity of the motion. The real motions were randomly picked from our filtered CMU Motion Capture Dataset. We distributed all of the 100 pairs in 10 Human Intelligence Tasks (HITs)\footnote{A Human Intelligence Task (HIT) is a task sent out and completed by one worker.}.  Each worker is allowed to do at most four HITs. Each HIT contains 10 testing pairs and 3 validation pairs. The validation pair contains a real motion sequence and a motion sequence with artificially large noise, so as to test if the worker is paying attention to the task. We only collected data from finished HITs that have correct answers for all three validation pairs. Ideally, if the synthetic motions have the same quality as the real motions, the workers will not be able to tell the real ones from fake ones, and the rate for picking up the synthetic ones out of all motion sequences should be close to 50\%. 

As listed in Table \ref{tab:user_study_results} under the train and test settings ``All-Random," the results show that around 40\% of our synthetic motions are preferred by the human workers, which indicates high quality. More results can be found in the supplemental video.

\subsection{Extension and Comparison with RTN}
\label{sect:rtn_comparison}
The related work Recurrent Transition Networks (RTN)~\cite{Harvey:2018:Recurrent} synthesizes motion transitions in a different setting than ours. Based on a fixed number of past frames and two future frames, RTN predicts a short sequence of future motion of a fixed length. Our network on the other hand supports arbitrary keyframe arrangement. In this section, we are going to examine how RTN works for short-duration motions from a single class, and how it works for long-duration motions mixed from multiple classes. We will also show the performance of our network trained under our keyframe arrangement setting, but tested under the RTN setting. 

\begin{figure}
    \includegraphics[width=0.9\linewidth]{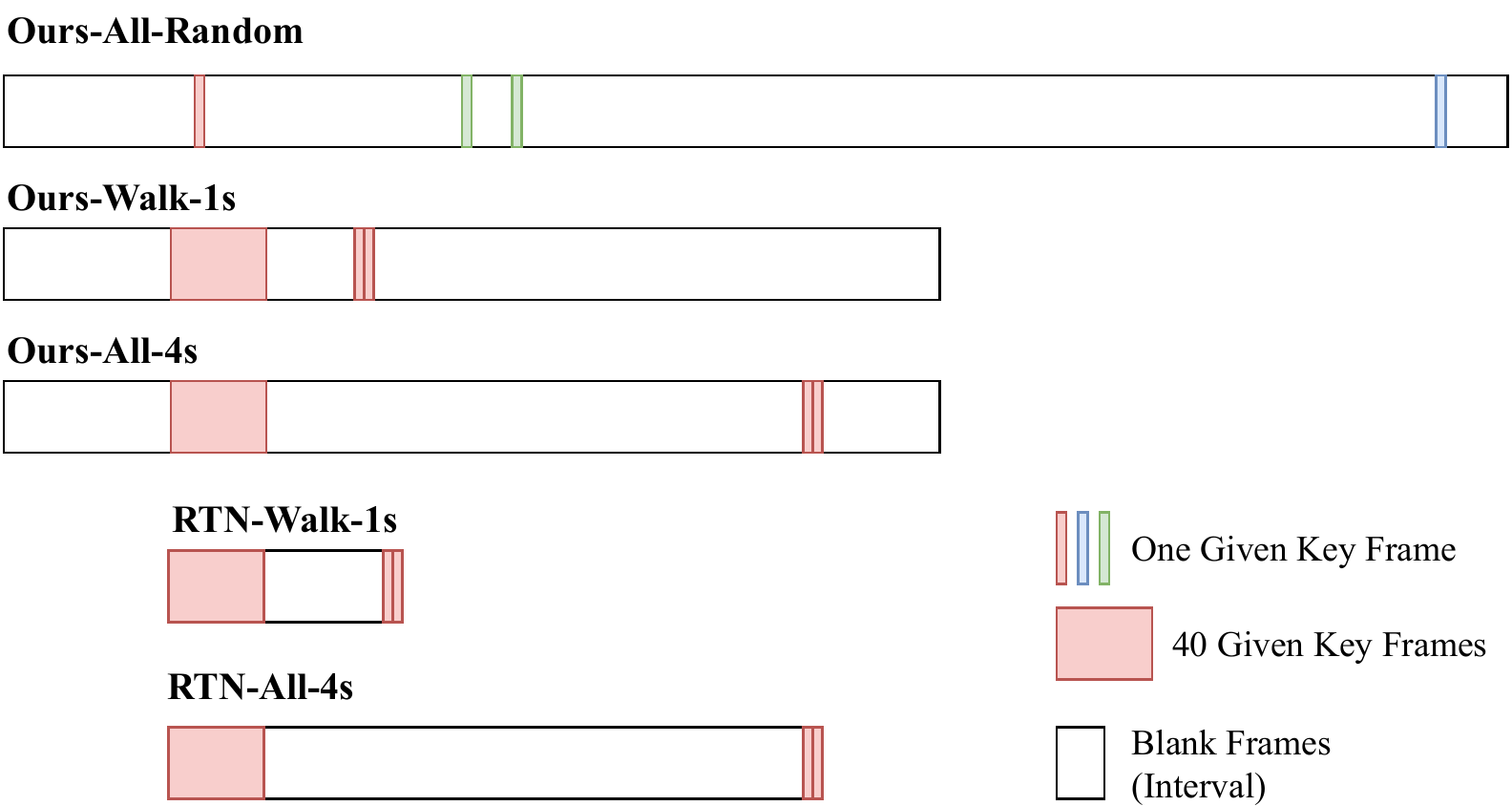}
    \caption{Training and testing settings for different approaches. These are defined in \sect{sect:rtn_comparison}.}
    \label{fig:comparison_testSetting}
\end{figure}

As illustrated in the first row of Figure~\ref{fig:comparison_testSetting}, our network is designed to generate various outputs with arbitrary length and arbitrary numbers of keyframes. It is trained and tested with sparsely sampled keyframes from 79 classes of motions that contain transitions within or across classes of poses: this scenario is labelled as ``Ours-All-Random" in the experiment (Table \ref{tab:user_study_results} and Figure~\ref{fig:comparison_testSetting}). Comparatively, RTN, in the original paper, is trained to predict 1 second of walking motion. In our experiment, we also trained and tested RTN's performance for predicting 1 second of walking motions, which we illustrate in Figure~\ref{fig:comparison_testSetting} as "RTN-Walk-1s." Moreover, we tested its performance for predicting longer (4 seconds) motions trained with all 79 classes: this is labelled as "RTN-All-4s". 

Even though our model is always trained on all 79 classes of the CMU dataset (per Section~ \ref{sec:train_details}), we also test our model's generalization abilities by testing in the same scenarios as RTN. The "Ours-Walk-1s" test, as a counterpart to "RTN-Walk-1s", uses one-second motion transitions with keyframes and DNA poses all sampled from the walking class only. Likewise, the "Ours-All-4s" test, as a counterpart to "RTN-All-4s", uses four-second motion transitions with keyframes and DNA poses sampled from all classes.

\input{table/user_study_result.tex}

To evaluate the quality of the synthetic motions, we generated 100 1-second sequences from "RTN-Walk-1s" and "Ours-Walk-1s," and 100 4-second sequences from "RTN-All-4s" and "Ours-All-4s" and performed the same user study on these sequences as is described in Section \ref{sec:motion_quality}. We also evaluated the keyframe alignment errors for each method based on 200 testing sequences for each method. As shown in Table \ref{tab:user_study_results}, RTN performs well in the "RTN-Walk-1s" case. However, for the 4 second case, RTN fails to generate reasonable results, with user preference rate  dropping to 2.8\% and Joint Error and Root Error increasing to 30.3 cm and 155.4 cm, respectively. On the other hand, our network, although not trained for the RTN tasks, still has reasonable user preference rate with low keyframe alignment error for both tasks. This indicates that our method not only performs well in the original training scenario but also generalizes well to new and unseen test scenarios. The user preference rates indicate that "Ours-All-4s" has higher motion quality than "Ours-All-1s." We believe this is because during training, the interval length is between 0 to 10 seconds, so the 4-second interval is closer to the expectation of the interval length during training.

\begin{figure}
    \includegraphics[width=0.8\linewidth]{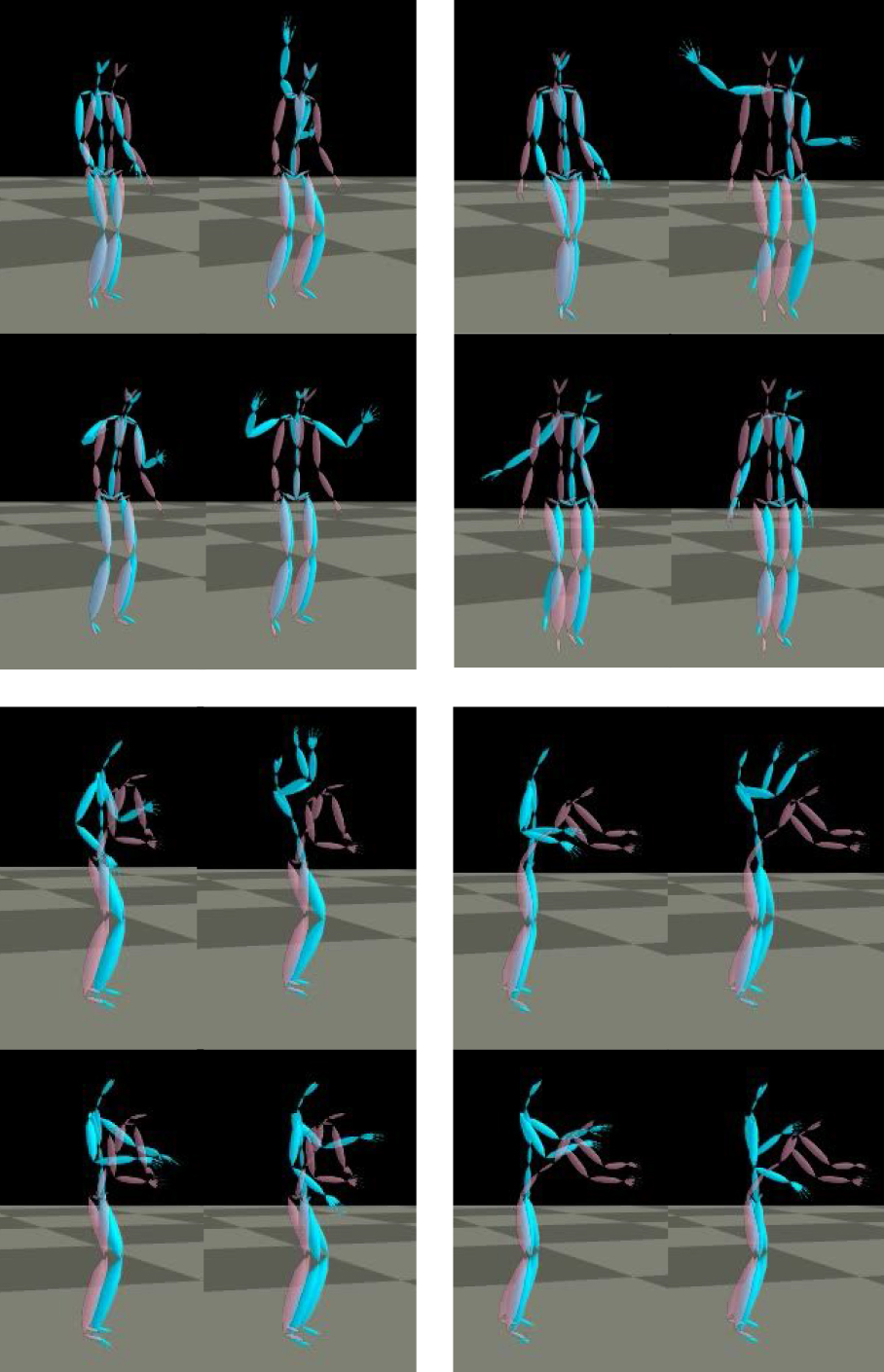}
    \caption{Examples of motion variety. Within each of the four sub-figures, we visualize four poses (blue) at the same frame from a motion sequence synthesized from the same set of keyframes but with four different Motion DNAs. The pink skeletons are the target keyframe pose in the near future.}
    \label{fig:four_poses}
\end{figure}

\begin{figure*}
    \includegraphics[width=\textwidth]{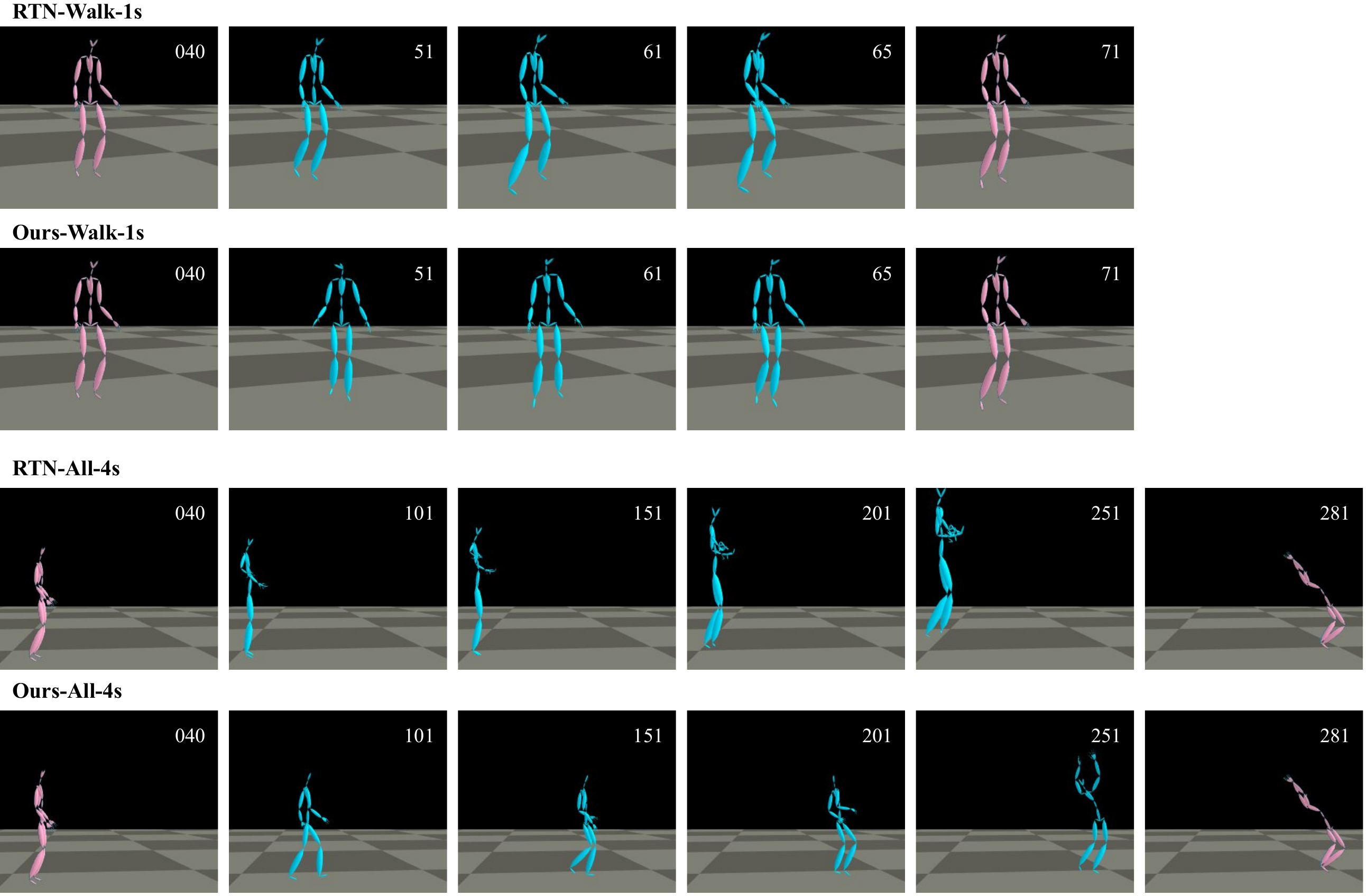}
    \caption{Examples of generating one second and four second transitions given the 40 past frames and 2 future frames. The first two rows show selected frames of a one-second transition sampled at 30 fps. The last two rows show frames of a four-second transition sampled at 60 fps. Pink skeletons visualize the input keyframes. From left to right, the blue skeletons show how the synthesized motion transitions between two keyframes. Numbers at the top-right corners are the frame indices. Corresponding results can be found in the supplementary video.}
    \label{fig:rtn_comparison_results}
\end{figure*}

In Figure \ref{fig:rtn_comparison_results}, we visualize qualitatively results of each method. In the first two lines, both RTN-Walk-1s model and our model is given the same past 40 frames and the future 2 frames, and they all successfully generate plausible walking transitions for the one second interval. In the last two lines, given the same input, our model generates a realistic transition while the RTN-All-4s model generates a diverging motion.

\subsection{Variation Control with Motion DNA}
To quantitatively evaluate the effect of our Motion DNA, we computed the two DNA errors as defined in Equation \ref{eqt:DNA1} and \ref{eqt:DNA2} of the 200 synthetic motion sequences described in Section \ref{sec:exp_keyframe_alignment}. The first DNA error is 8.4 cm, which means all of the representative poses in the motion DNA can be found somewhere in the synthesized motion sequence. The second DNA error is 6.5 cm, which means these representative poses are distributed evenly in the results. 

\begin{figure*}
    \includegraphics[width=0.95\textwidth]{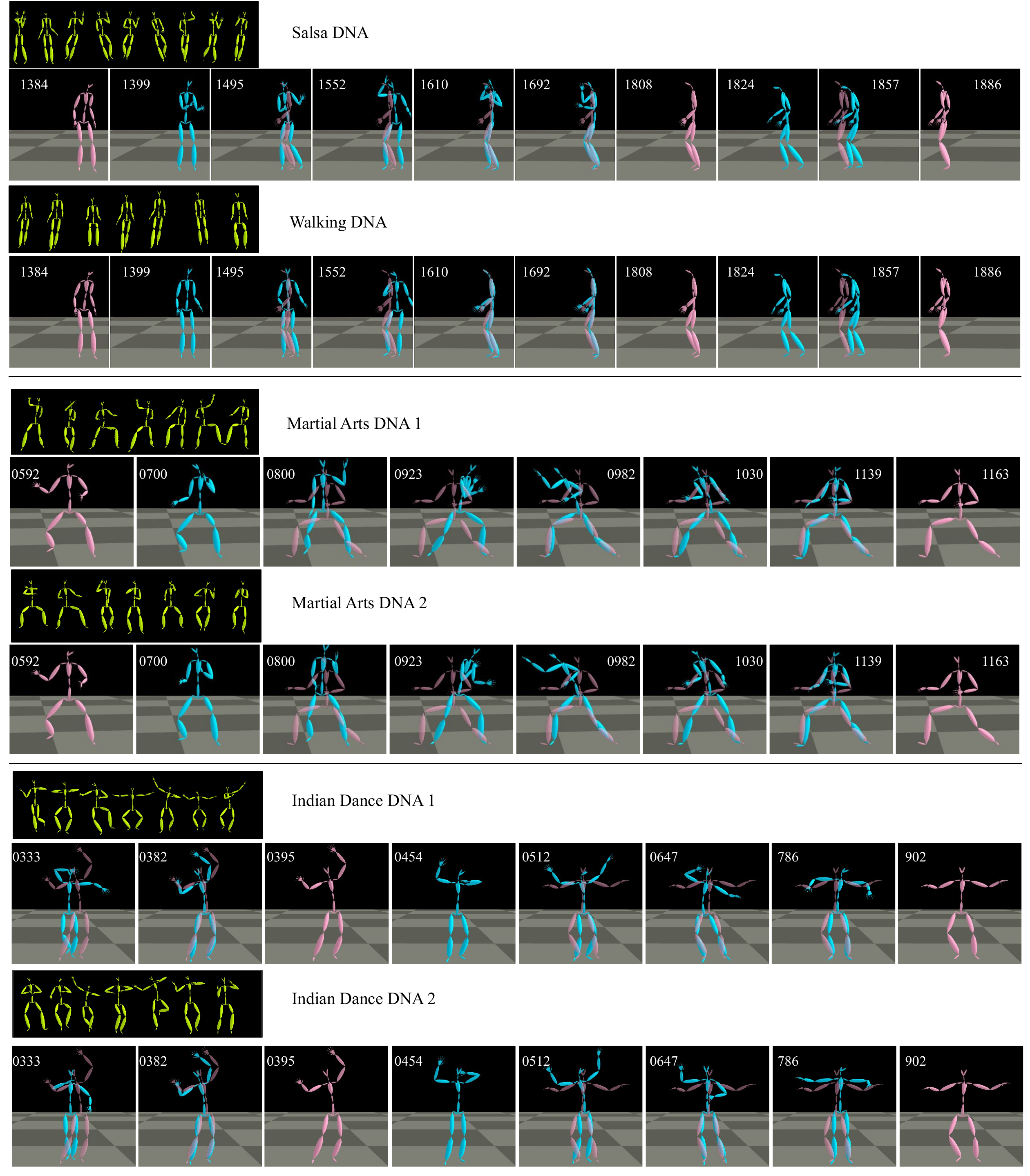}
    \caption{Examples of inbetweening. The pink skeletons visualize the user-specified keyframes. From left to right, the blue skeletons show how the synthesized motion transitions between two keyframes. The semi-transparent pink skeletons are the keyframes in the near future within 400 frames. Each group of two rows shows frames from two generated 2048-frame-long synthetic motion sequences given the same input keyframes but different Motion DNA. The yellow skeletons are the representative poses for the Motion DNA.}
    \label{fig:DNA_effect}
\end{figure*}

We also visualize the effect of Motion DNA in Figure \ref{fig:DNA_effect}. 
In the top two rows, we use motion DNA extracted from salsa poses (top row) and walking poses (second from the top) as input to the network. Under the same keyframe constraints (transparent pink skeletons), the network generates salsa (raised arms in the top row) and walking styles respectively. 
In the middle two rows, we use two different martial arts motion DNAs, and both results resemble martial arts but exhibit some differences in the gestures of the arms. 
A similar observation applies to the bottom two rows, where Indian Dance DNAs are used as the seed of motion. 
Figure~\ref{fig:teaser} shows another example. Under the same keyframe constraints, the synthesized two motions have different trajectories and local motions, but they both travel to the same spots at the keyframes.
Figure \ref{fig:four_poses} gives more variation examples. We show the same frame of a synthesized sequence which is generated from the same keyframes but with four different Motion DNAs. 
We observe that variations happen more frequently in the upper body than the lower body. Since the foot motion has more impact on the global movement, it is harder to find foot patterns that exactly satisfy the keyframe constraints. 

%% file: table/eval_representation.tex
\begin{table}[]
\begin{tabular}{|l|l|l|l|}
\hline
                                                                        & Mean(cm) & Max(cm) & Std(cm) \\ \hline
Quaternion                                                                    & 9.03     & 179.66  & 16.33   \\ \hline
Vanilla Euler Angles                                                          & 2.7      & 48.7    & 2.1     \\ \hline
\begin{tabular}[c]{@{}l@{}}Range-Constrained\\ Euler + 6D root (Ours)\end{tabular} & 2.3      & 34.9    & 1.5     \\ \hline
\end{tabular}
\caption{IK Reconstruction errors using different rotation representations. The quaternion and vanilla Euler angle representations are used for all nodes. Ours uses the range-constrained Euler representation for non-root nodes and the continuous 6D representation~\cite{zhou2019} for the root node.}
\label{tab:RP}
\end{table}

%% file: table/eval_path_prediction.tex
\begin{table}[]
\begin{tabular}{|l|l|l|l|l|l|l|l|l|}
\hline
$V_{1}$ & $V_{2}$ & $V_{4}$ & $V_{8}$ & $V_{16}$ & $V_{32}$ & $V_{64}$ & $V_{128}$ & Y   \\ \hline
0.7  & 0.8  & 1.1  & 1.5  & 1.9   & 2.8   & 3.9   & 5.5    & 1.8 \\ \hline
\end{tabular}
\caption{Global path prediction mean errors in centimeters. $V_n$ is the mean error of the root (hip) translation differences in the x-z plane for poses predicted at $n$ frames in the future. $Y$ is the mean error of the root (hip) along the y axis.}
\label{tab:hip_prediction_error}
\end{table}

%% file: table/eval_speed.tex
\begin{table}[]
\begin{tabular}{|l|l|l|l|l|}
\hline
Sequence Length (frames) & \begin{tabular}[c]{@{}l@{}}512\\ (8 s)\end{tabular} & \begin{tabular}[c]{@{}l@{}}1024\\ (17 s)\end{tabular} & \begin{tabular}[c]{@{}l@{}}2048\\ (34 s)\end{tabular} & \begin{tabular}[c]{@{}l@{}}4096\\ (68 s)\end{tabular} \\ \hline
Local Motion Generation (s)      & 0.019       & 0.023         & 0.021                & 0.022              \\ \hline
Global Path Prediction (s)      & 0.033       & 0.061          & 0.131                 & 0.237              \\ \hline
Post-processing (s)       & 0.008             & 0.017          & 0.036                 & 0.070                                                  \\ \hline
Total (s)                & 0.059             & 0.101         & 0.188                  & 0.329                                                  \\ \hline
\end{tabular}
\caption{Mean Computation Time for Generating Different Lengths of Motions. s refers to second.  }
\label{tab:speed}
\end{table}

%% file: table/user_study_result.tex
\begin{table}[]
\small
\begin{tabular}{|l|c|c|c|c|c|}
\hline
\multicolumn{6}{|c|}{User Study}                                                                                                                     \\ \hline
                                                                 & \multicolumn{3}{c|}{\textbf{Ours}} & \multicolumn{2}{c|}{\textbf{RTN}} \\ \hline
Train Setting                                                    & \multicolumn{3}{c|}{All-Random}               & Walk-1s          & All-4s         \\ \hline
Test Setting                                                     & All-Random       & Walk-1s      & All-4s      & Walk-1s          & All-4s         \\ \hline
Real                                                      & 193              & 216          & 217         & 156              & 314            \\ \hline
Synthetic                                                 & 124              & 94           & 163         & 154              & 9              \\ \hline
\begin{tabular}[c]{@{}l@{}}User Preference \end{tabular} & 39.1\%           & 30.3\%       & 42.9\%      & 49\%             & 2.8\%          \\ \hline
Margin of Error                                                  & $\pm$5.4\%           & $\pm$5.1\%       & $\pm$5.0\%      & $\pm$5.6\%           & $\pm$1.8\%         \\ \hline
\multicolumn{6}{|c|}{Input frame Alignment Error}                                                                                                       \\ \hline
Joint Error (cm)                                            & 3.5              & 4.4          & 5.6         & 2.3              & 30.3           \\ \hline
Root Error (cm)                                             & 10.0             & 5.1          & 4.5         & 10.7             & 155.4          \\ \hline
\end{tabular}
\caption{Evaluation results for motion quality and keyframe alignment. The first four rows of numbers are the user study results collected from 101 human workers. The rows ``real" and ``synthetic" are the number of workers who chose the real or synthetic motions, respectively. ``User preference" is the percentage of synthetic motions  chosen out of total pairs, and the margins of error are listed on the next row with confidence level at 95\%. The last two rows are the mean Euclidean error of the global root positions and the local joint positions of the input keyframes. Please refer to \sect{sect:rtn_comparison} for the definitions of the training and test settings.}
\label{tab:user_study_results}
\end{table}

%% file: Limitation.tex
\subsection{Limitations}
Although our network can interpolate keyframe poses over a much longer time interval than traditional methods, to guarantee the naturalness of the synthesized transition, the maximum allowable time interval between any neighbouring keyframes is still limited by the size of the receptive field of the generator. A possible solution to support even longer time intervals is to do hierarchical synthesis which synthesizes some critical frames first and then generates motions in shorter intervals. Our network operates at a fixed temporal resolution. If two completely different keyframes are specified in a very short interval and their transition is not achievable at the given speed, then one of them may be filtered out as an outlier. 

Our Motion DNA scheme provides a soft control on the style of the output motion. How much it impacts the output style depends on the keyframe constraints. For example, as shown in the supplemental video, when the keyframes and the Motion DNA are both poses from martial arts, the output usually capture the characteristics of martial arts in both arms and legs. However, when the keyframes are walking poses, the result will look less like martial arts, where the output has punching and defending gestures in the upper body but with walking pose in the lower body. Another limitation is that because it uses static representative frames, the Motion DNA contains only the iconic postures of the motion, and might fail to capture the style reflected in the subtle body dynamics. Additionally, in some cases, such as Case 2 in the video, the motions near keyframes can look similar even though different Motion DNAs were provided.

Our method can at times produce foot sliding or skating artifacts, or feet may not properly contact the ground. As we demonstrate in the video for the rigged animation result, these can be fixed by a simple  postprocessing~\cite{li2017auto}. In future work, a loss could be adopted to minimize foot sliding or skating, similar to Holden~et~al.~\shortcite{holden2016deep}: we believe this may be tricky for the CMU dataset we used because there are some dance motions that intentionally use foot sliding.

Some of the generated motions from our method can have overly fast root rotation, such as Case 2 in the video, or be a bit monotonous, such as Case 4 in the video. Future work might address these issues by for example using additional losses.

%% file: additional_features.tex
\begin{figure*}
    \includegraphics[width=0.85\textwidth]{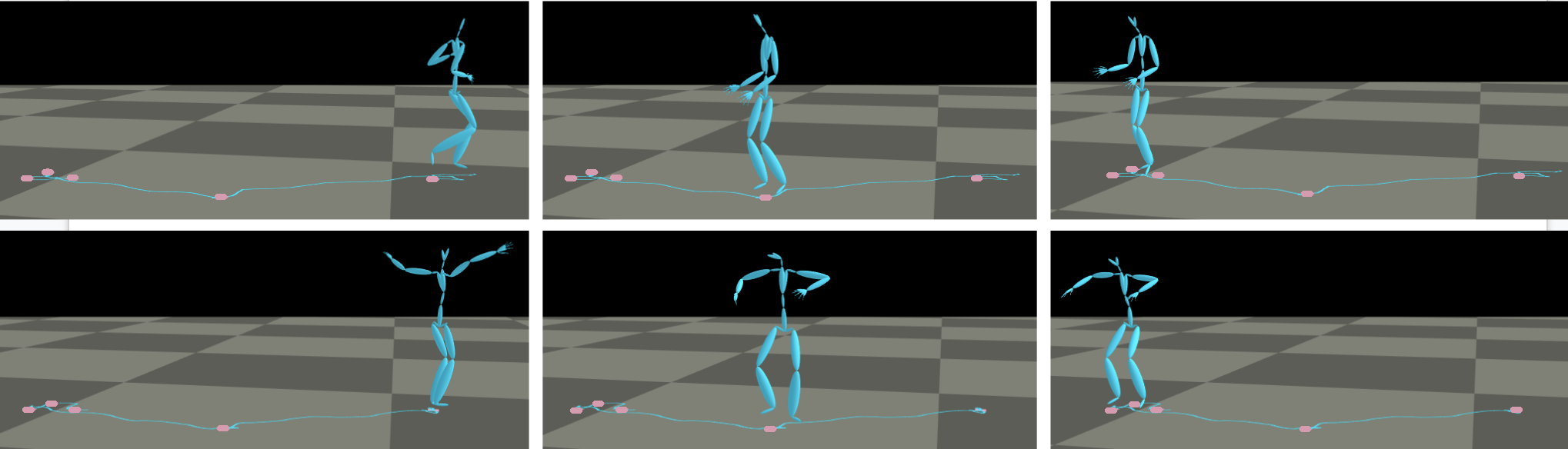}
    \caption{Motion Generation given sparse root coordinates. Two synthesized motion sequences given the same keyframes (position of the root joints as pink dots on the ground) and different representative frames. Top row: the $100$th, $500$th, and $700$th frame from a synthesized sequence. Bottom row: the corresponding frames from another sequence synthesized with a different set of representative frames.}
    \label{fig:root}
\end{figure*}

 \begin{figure*}
    \includegraphics[width=0.6\textwidth]{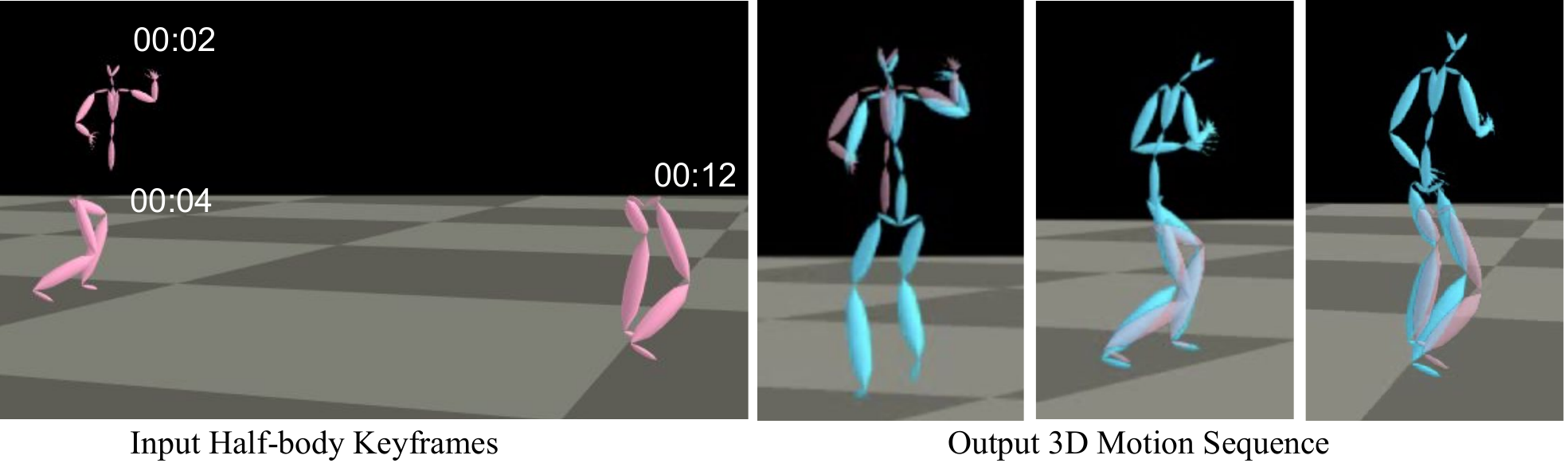}
    \caption{Motion Generation given random half-body joints. Left: keyframe partial joints inputs and their times. Right: the synthesized 3D poses at keyframes (pink) and nearby frames (blue). }
    \label{fig:halfbody}
\end{figure*}

 \begin{figure*}
    \includegraphics[width=0.85\textwidth]{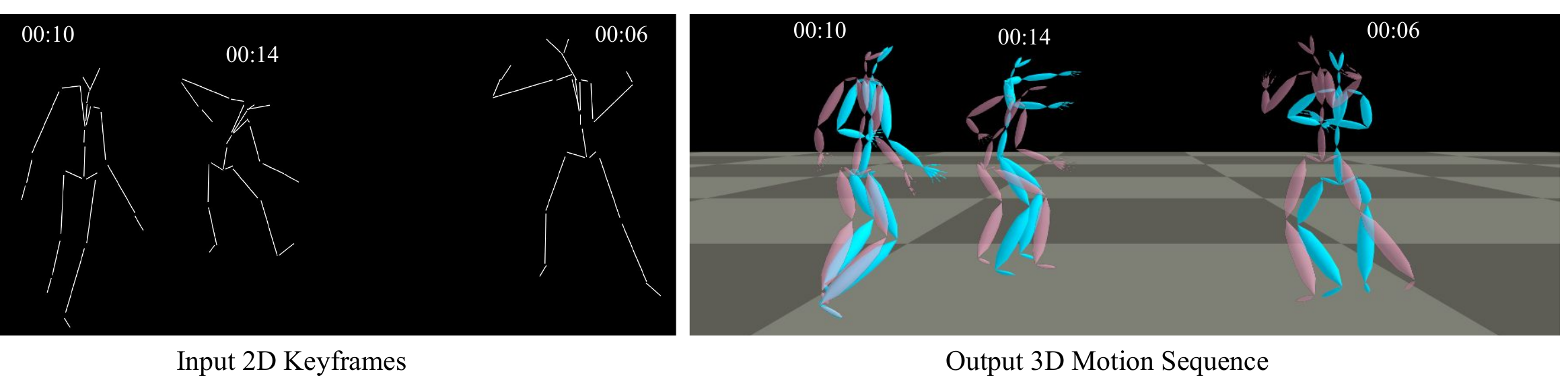}
    \caption{Motion Generation given manually made 2D poses. Left: 2D keyframe inputs drawn on the x-y plane and their times. Right: the synthesized 3D poses at keyframes (pink) and nearby frames (blue). }
    \label{fig:2D}
\end{figure*}
\section{Additional Features}
In addition to the features discussed above, our method can also be used when only part of the body pose is specified at keyframes. We can change the format of the input (Section \ref{sec:input_sequence_format}) to the network, so that coordinates are only specified at the desired joints and the rest of the joints are masked out in the input matrix during both, training and testing.
\subsection{Partial Body Control: Root}
Figure~\ref{fig:root} shows the case where only the root joint positions are specified by the user at sparse locations (pink dots on the ground). The two rows show two synthesized sequences generated with the same root joint positions but different representative frames. From left to right, the three frames are the $100$th, $500$th, and $700$th frames. Note that the two generated sequences exhibit different root trajectories and local motions.

\subsection{Partial Body Control: Half-body} 
We trained a network that the keyframe input can contain either the whole body joints or only half of the body's joints. Figure~\ref{fig:halfbody} shows the 3D motion sequences synthesized from partially given keyframe poses.

\subsection{Partial Body Control: 2D Joints} 
Users might want to create 3D skeleton poses by sketching the keyframes in 2D. For this task, we trained a network by providing only the x and y coordinates of the joints and the root. To handle non-standard pose input such as casually drawn stick figures, we trained the network with augmented 2D poses. In Figure \ref{fig:2D}, we show that our network is able to lift the 2D keyframes into 3D and synthesize natural 3D motions. Although the input has stretched and uneven bone lengths, the network manage to synthesize feasible 3D poses that mimic the 2D input. The recent work VideoPose3D~\cite{pavllo:videopose3d:2019} also predicts 3D motion from the 2D motion, but it cannot perform the long-term motion inbetweening task.

%% file: conclusion.tex
\section{Conclusion and Future Work}
We proposed a novel deep generative modeling approach, that can synthesize extended and complex natural motions between highly sparse keyframes of 3D human body poses. Our key insight is that conditional GANs with large receptive field and appropriate input representation are suitable for enforcing keyframe constraints in a holistic way. However, although GANs can generate realistic local motions, they tend to diverge during training for generating global motions. From extensive experiments, we discovered that global motions can be predicted from local motions using neural networks. Thus, we have introduced a two-stage generative adversarial network that first synthesizes local joint rotations and then predicts a global path for the character to follow. 

We also proposed a novel range-constrained forward kinematics layer that helps the network explore a continuous rotation representation space with biomechanical constraints and thus, can produce natural-looking motions. To synthesize different results with the same keyframes, we introduced a new authoring concept, called Motion DNA, which allows users to influence the style of the synthesized output motion.

We compared our method with a closely related work, RTN~\cite{Harvey:2018:Recurrent} and showed superior results for our task. Not only does our method support additional capabilities, our trained deep model also generalizes to RTN's task. Our method handles more complex motions and provides users the flexibility of choosing their preferred sequence among a variety of synthesized results.

In the near future, we plan to train a more flexible system that allows users to control various subsets of joints or specify an arbitrary trajectory. This can be easily achieved by masking out random joints or providing trajectories in the input during training, which is supported by our input format. Another interesting direction is to extend our method to long-term inbetweening of non-human characters. While our current approach is highly efficient and can perform in real-time for short sequences, being able to update extended motion sequences in real-time would enable new possibilities and impact next-generation character animation control systems in gaming and motion planning in robotics.


%% file: appendix.tex
\newpage
\appendix


\section{Training Parameters}
\label{sec:train_param}
For training the inbetweening network, we use learning rate 0.00001 for both the generator and the discriminator. The weight for the Least Square GAN loss is 1.0, 5.0 for the batch normalization loss, 300 for local joints error, 50 for each DNA loss. The weight for the global path loss increases by 1.0 from 10.0 to 80.0 every 1000 iterations. 
Table \ref{tab:network_architectures} gives the parameters of the network architecture. Table \ref{tab:rotation} gives the range and orders of relative rotations at each joint.

\section{Dataset Filtering}
The original CMU Motion Capture dataset contains many noises. There are three types of noises: (1) motions underneath or cutting through the ground surface, (2) motions with poses outside the range of human body flexibility, (3) jittering due to the limitation of the capture device. We first remove the first type by calculating the average lowest position of the motion sequences over a short time window. For detecting the second and third type of noises, we trained a local motion auto-encoder with similar architecture to the local motion generator, but with less layers. The auto-encoder was trained with all the motion classes we use in this paper. Since the auto-encoder contains the RC-FC layer, the output poses are all under the rotation range constraints of human joints. Moreover, as anto-encoder can filter out high-frequency signals, the output motions are usually smoother than the input motions. We apply the trained auto-encoder to all the motion sequences and compute the error between the input frames and output frames. When the error is higher than a threshold, we delete the corresponding frame and split the motion clip at that frame. 

\section{Keyframe Sampling Scheme}
At training time, we sample keyframes in a way that plausibly emulates how an animator might place keyframes both in temporal spacing and in similarity of their pose. This allows our trained network to generalize to different possible keyframe configurations that an animator might give at test time. As explained in Section~\ref{sec:generator_and_discriminator}, to synthesize natural transitions, the longest allowable time interval between keyframes is 636 frames. Therefore, we first randomly pick the length of the intervals to be between 0 and 600 frames (10 seconds for 60 fps and 20 seconds for 30 fps). 
Based on the sampled interval lengths, we sample the keyframes from the beginning of the motion sequence to the end, based on two rules: 
(1) When the current interval is shorter than three seconds, we sample the two keyframes from the same clip so that their temporal order is maintained and the number of frames between their locations in the ground truth clip equals the current interval length. This is because it is hard for the network to improvise smooth transitions in a short time interval if sampled keyframes have dramatically different pose due to them coming from different motion clips or being far apart in the same clip. (2) When the current interval exceeds three seconds, we want to simulate the scenario of transitioning between different motion clips of the same or different classes. In such circumstance, with a certain probability, we sample the next keyframe from a different motion clip other than the clip where the last keyframe comes from, and place it at a random position. The probability and the root distance between the two keyframes are proportional to the current interval length. The first row of Figure~\ref{fig:comparison_testSetting} gives a keyframe placement example.

\section{Representative Frames for Motion DNA Sampling}
For training the Motion DNA encoder, the number of representative frames we use as input to the network is proportional to the length of the interval between neighboring keyframes. Specifically, we use 1 representative frame for every 3-second-long interval. 

\input{table/networks.tex}
\input{table/rotation_range.tex}

%% file: table/networks.tex
\begin{table*}[]
\small\addtolength{\tabcolsep}{-1pt}
\begin{tabular}{llllllll}
\multicolumn{8}{l}{Encoder}                                                                                                                                                                                                                             \\ \hline
\multicolumn{1}{|l|}{Layer}      & \multicolumn{1}{l|}{in\_ch} & \multicolumn{1}{l|}{out\_ch}  & \multicolumn{1}{l|}{kernel size} & \multicolumn{1}{l|}{stride} & \multicolumn{1}{l|}{pad} & \multicolumn{1}{l|}{ratio} & \multicolumn{1}{l|}{out\_len} \\ \hline
\multicolumn{1}{|l|}{res\_1}     & \multicolumn{1}{l|}{3M+3M}  & \multicolumn{1}{l|}{384}      & \multicolumn{1}{l|}{4}      & \multicolumn{1}{l|}{2}      & \multicolumn{1}{l|}{1}   & \multicolumn{1}{l|}{1/1}        & \multicolumn{1}{l|}{N/2}      \\ \hline
\multicolumn{1}{|l|}{res\_2}     & \multicolumn{1}{l|}{384}    & \multicolumn{1}{l|}{384}      & \multicolumn{1}{l|}{4}      & \multicolumn{1}{l|}{2}      & \multicolumn{1}{l|}{1}   & \multicolumn{1}{l|}{1/2}        & \multicolumn{1}{l|}{N/4}      \\ \hline
\multicolumn{1}{|l|}{res\_3}     & \multicolumn{1}{l|}{384}    & \multicolumn{1}{l|}{512}      & \multicolumn{1}{l|}{4}      & \multicolumn{1}{l|}{2}      & \multicolumn{1}{l|}{1}   & \multicolumn{1}{l|}{1/3}        & \multicolumn{1}{l|}{N/8}      \\ \hline
\multicolumn{1}{|l|}{res\_4}     & \multicolumn{1}{l|}{512}    & \multicolumn{1}{l|}{512}      & \multicolumn{1}{l|}{4}      & \multicolumn{1}{l|}{2}      & \multicolumn{1}{l|}{1}   & \multicolumn{1}{l|}{1/4}        & \multicolumn{1}{l|}{N/16}     \\ \hline
\multicolumn{1}{|l|}{res\_5}     & \multicolumn{1}{l|}{512}    & \multicolumn{1}{l|}{768}      & \multicolumn{1}{l|}{4}      & \multicolumn{1}{l|}{2}      & \multicolumn{1}{l|}{1}   & \multicolumn{1}{l|}{1/5}        & \multicolumn{1}{l|}{N/32}     \\ \hline
\multicolumn{1}{|l|}{res\_6}     & \multicolumn{1}{l|}{768}    & \multicolumn{1}{l|}{1024}     & \multicolumn{1}{l|}{4}      & \multicolumn{1}{l|}{2}      & \multicolumn{1}{l|}{1}   & \multicolumn{1}{l|}{1/6}        & \multicolumn{1}{l|}{N/64}     \\ \hline
\multicolumn{8}{l}{Motion DNA Encoder}                                                                                                                                                                                                                  \\ \hline
\multicolumn{1}{|l|}{conv\_1}    & \multicolumn{1}{l|}{3(M-1)} & \multicolumn{1}{l|}{1024}     & \multicolumn{1}{l|}{1}      & \multicolumn{1}{l|}{1}      & \multicolumn{1}{l|}{0}   & \multicolumn{1}{l|}{-}          & \multicolumn{1}{l|}{$\hat{N}$}        \\ \hline
\multicolumn{1}{|l|}{affine} & \multicolumn{1}{l|}{1024}   & \multicolumn{1}{l|}{1024}     & \multicolumn{1}{l|}{}       & \multicolumn{1}{l|}{}       & \multicolumn{1}{l|}{}    & \multicolumn{1}{l|}{}           & \multicolumn{1}{l|}{$\hat{N}$}        \\ \hline
\multicolumn{1}{|l|}{CPReLU}     & \multicolumn{1}{l|}{1024}   & \multicolumn{1}{l|}{1024}     & \multicolumn{1}{l|}{}       & \multicolumn{1}{l|}{}       & \multicolumn{1}{l|}{}    & \multicolumn{1}{l|}{}           & \multicolumn{1}{l|}{$\hat{N}$}        \\ \hline
\multicolumn{1}{|l|}{conv\_2}    & \multicolumn{1}{l|}{1024}   & \multicolumn{1}{l|}{1024}     & \multicolumn{1}{l|}{}       & \multicolumn{1}{l|}{}       & \multicolumn{1}{l|}{}    & \multicolumn{1}{l|}{}           & \multicolumn{1}{l|}{$\hat{N}$}        \\ \hline
\multicolumn{1}{|l|}{affine} & \multicolumn{1}{l|}{1024}   & \multicolumn{1}{l|}{1024}     & \multicolumn{1}{l|}{}       & \multicolumn{1}{l|}{}       & \multicolumn{1}{l|}{}    & \multicolumn{1}{l|}{}           & \multicolumn{1}{l|}{$\hat{N}$}        \\ \hline
\multicolumn{1}{|l|}{AvgPool}    & \multicolumn{1}{l|}{1024}   & \multicolumn{1}{l|}{1024}     & \multicolumn{1}{l|}{}       & \multicolumn{1}{l|}{}       & \multicolumn{1}{l|}{}    & \multicolumn{1}{l|}{}           & \multicolumn{1}{l|}{1}        \\ \hline
\multicolumn{8}{l}{Decoder}                                                                                                                                                                                                                             \\ \hline
\multicolumn{1}{|l|}{res\_1}     & \multicolumn{1}{l|}{2048}   & \multicolumn{1}{l|}{1024}     & \multicolumn{1}{l|}{1}      & \multicolumn{1}{l|}{1}      & \multicolumn{1}{l|}{0}   & \multicolumn{1}{l|}{1/1}        & \multicolumn{1}{l|}{N/64}     \\ \hline
\multicolumn{1}{|l|}{res\_2}     & \multicolumn{1}{l|}{1024}   & \multicolumn{1}{l|}{1024}     & \multicolumn{1}{l|}{3}      & \multicolumn{1}{l|}{1}      & \multicolumn{1}{l|}{1}   & \multicolumn{1}{l|}{1/1.4}      & \multicolumn{1}{l|}{N/64}     \\ \hline
\multicolumn{1}{|l|}{res\_t\_3}  & \multicolumn{1}{l|}{1024}   & \multicolumn{1}{l|}{1024}     & \multicolumn{1}{l|}{4}      & \multicolumn{1}{l|}{2}      & \multicolumn{1}{l|}{1}   & \multicolumn{1}{l|}{1/1.6}      & \multicolumn{1}{l|}{N/32}     \\ \hline
\multicolumn{1}{|l|}{res\_4}     & \multicolumn{1}{l|}{1024}   & \multicolumn{1}{l|}{768}      & \multicolumn{1}{l|}{3}      & \multicolumn{1}{l|}{1}      & \multicolumn{1}{l|}{1}   & \multicolumn{1}{l|}{1/2.2}      & \multicolumn{1}{l|}{N/32}     \\ \hline
\multicolumn{1}{|l|}{res\_t\_5}  & \multicolumn{1}{l|}{768}    & \multicolumn{1}{l|}{768}      & \multicolumn{1}{l|}{4}      & \multicolumn{1}{l|}{2}      & \multicolumn{1}{l|}{1}   & \multicolumn{1}{l|}{1.0/2.8}    & \multicolumn{1}{l|}{N/16}     \\ \hline
\multicolumn{1}{|l|}{res\_6}     & \multicolumn{1}{l|}{768}    & \multicolumn{1}{l|}{768}      & \multicolumn{1}{l|}{3}      & \multicolumn{1}{l|}{1}      & \multicolumn{1}{l|}{1}   & \multicolumn{1}{l|}{1.0/3.6}    & \multicolumn{1}{l|}{N/16}     \\ \hline
\multicolumn{1}{|l|}{res\_t\_7}  & \multicolumn{1}{l|}{768}    & \multicolumn{1}{l|}{768}      & \multicolumn{1}{l|}{4}      & \multicolumn{1}{l|}{2}      & \multicolumn{1}{l|}{1}   & \multicolumn{1}{l|}{1.0/4.6}    & \multicolumn{1}{l|}{N/8}      \\ \hline
\multicolumn{1}{|l|}{res\_8}     & \multicolumn{1}{l|}{768}    & \multicolumn{1}{l|}{512}      & \multicolumn{1}{l|}{3}      & \multicolumn{1}{l|}{1}      & \multicolumn{1}{l|}{1}   & \multicolumn{1}{l|}{1/5.8}      & \multicolumn{1}{l|}{N/8}      \\ \hline
\multicolumn{1}{|l|}{res\_t\_9}  & \multicolumn{1}{l|}{512}    & \multicolumn{1}{l|}{512}      & \multicolumn{1}{l|}{4}      & \multicolumn{1}{l|}{2}      & \multicolumn{1}{l|}{1}   & \multicolumn{1}{l|}{1/7.2}      & \multicolumn{1}{l|}{N/4}      \\ \hline
\multicolumn{1}{|l|}{res\_10}    & \multicolumn{1}{l|}{512}    & \multicolumn{1}{l|}{512}      & \multicolumn{1}{l|}{3}      & \multicolumn{1}{l|}{1}      & \multicolumn{1}{l|}{1}   & \multicolumn{1}{l|}{1/8.8}      & \multicolumn{1}{l|}{N/4}      \\ \hline
\multicolumn{1}{|l|}{res\_t\_11} & \multicolumn{1}{l|}{512}    & \multicolumn{1}{l|}{512}      & \multicolumn{1}{l|}{4}      & \multicolumn{1}{l|}{2}      & \multicolumn{1}{l|}{1}   & \multicolumn{1}{l|}{1/10.6}     & \multicolumn{1}{l|}{N/2}      \\ \hline
\multicolumn{1}{|l|}{res\_12}    & \multicolumn{1}{l|}{512}    & \multicolumn{1}{l|}{512}      & \multicolumn{1}{l|}{3}      & \multicolumn{1}{l|}{1}      & \multicolumn{1}{l|}{1}   & \multicolumn{1}{l|}{1/12.6}     & \multicolumn{1}{l|}{N/2}      \\ \hline
\multicolumn{1}{|l|}{res\_t\_13} & \multicolumn{1}{l|}{512}    & \multicolumn{1}{l|}{512}      & \multicolumn{1}{l|}{4}      & \multicolumn{1}{l|}{2}      & \multicolumn{1}{l|}{1}   & \multicolumn{1}{l|}{1/14.8}     & \multicolumn{1}{l|}{N}        \\ \hline
\multicolumn{1}{|l|}{res\_14}    & \multicolumn{1}{l|}{512}    & \multicolumn{1}{l|}{6+3(M-1)} & \multicolumn{1}{l|}{3}      & \multicolumn{1}{l|}{1}      & \multicolumn{1}{l|}{1}   & \multicolumn{1}{l|}{1/17.2}     & \multicolumn{1}{l|}{N}        \\ \hline
\multicolumn{8}{l}{Discriminator}                                                                                                                                                                                                                       \\ \hline
\multicolumn{1}{|l|}{res\_1}     & \multicolumn{1}{l|}{3(M-1)} & \multicolumn{1}{l|}{512}      & \multicolumn{1}{l|}{4}      & \multicolumn{1}{l|}{2}      & \multicolumn{1}{l|}{1}   & \multicolumn{1}{l|}{1/1}        & \multicolumn{1}{l|}{N/2}      \\ \hline
\multicolumn{1}{|l|}{res\_2}     & \multicolumn{1}{l|}{512}    & \multicolumn{1}{l|}{512}      & \multicolumn{1}{l|}{4}      & \multicolumn{1}{l|}{2}      & \multicolumn{1}{l|}{1}   & \multicolumn{1}{l|}{1/2}        & \multicolumn{1}{l|}{N/4}      \\ \hline
\multicolumn{1}{|l|}{res\_3}     & \multicolumn{1}{l|}{512}    & \multicolumn{1}{l|}{512}      & \multicolumn{1}{l|}{4}      & \multicolumn{1}{l|}{2}      & \multicolumn{1}{l|}{1}   & \multicolumn{1}{l|}{1/3}        & \multicolumn{1}{l|}{N/8}      \\ \hline
\multicolumn{1}{|l|}{res\_4}     & \multicolumn{1}{l|}{512}    & \multicolumn{1}{l|}{512}      & \multicolumn{1}{l|}{4}      & \multicolumn{1}{l|}{2}      & \multicolumn{1}{l|}{1}   & \multicolumn{1}{l|}{1/4}        & \multicolumn{1}{l|}{N/16}     \\ \hline
\multicolumn{1}{|l|}{res\_5}     & \multicolumn{1}{l|}{512}    & \multicolumn{1}{l|}{1024}     & \multicolumn{1}{l|}{4}      & \multicolumn{1}{l|}{2}      & \multicolumn{1}{l|}{1}   & \multicolumn{1}{l|}{1/5}        & \multicolumn{1}{l|}{N/32}     \\ \hline
\multicolumn{1}{|l|}{res\_6}     & \multicolumn{1}{l|}{1024}   & \multicolumn{1}{l|}{1024}     & \multicolumn{1}{l|}{4}      & \multicolumn{1}{l|}{2}      & \multicolumn{1}{l|}{1}   & \multicolumn{1}{l|}{1/6}        & \multicolumn{1}{l|}{N/64}     \\ \hline
\multicolumn{1}{|l|}{conv\_7}    & \multicolumn{1}{l|}{1024}   & \multicolumn{1}{l|}{1}        & \multicolumn{1}{l|}{1}      & \multicolumn{1}{l|}{1}      & \multicolumn{1}{l|}{0}   & \multicolumn{1}{l|}{}           & \multicolumn{1}{l|}{N/64}     \\ \hline

\multicolumn{8}{l}{Global Path Predictor}                                                                                                                                                                                                                       \\ \hline
\multicolumn{1}{|l|}{res\_1}     & \multicolumn{1}{l|}{3(M-1)} & \multicolumn{1}{l|}{128}      & \multicolumn{1}{l|}{12}      & \multicolumn{1}{l|}{2}      & \multicolumn{1}{l|}{5}   & \multicolumn{1}{l|}{1/1}        & \multicolumn{1}{l|}{N/2}      \\ \hline
\multicolumn{1}{|l|}{res\_2}     & \multicolumn{1}{l|}{128}    & \multicolumn{1}{l|}{128}      & \multicolumn{1}{l|}{12}      & \multicolumn{1}{l|}{2}      & \multicolumn{1}{l|}{5}   & \multicolumn{1}{l|}{1/2}        & \multicolumn{1}{l|}{N/4}      \\ \hline
\multicolumn{1}{|l|}{res\_3}     & \multicolumn{1}{l|}{128}    & \multicolumn{1}{l|}{256}      & \multicolumn{1}{l|}{12}      & \multicolumn{1}{l|}{2}      & \multicolumn{1}{l|}{5}   & \multicolumn{1}{l|}{1/3}        & \multicolumn{1}{l|}{N/8}      \\ \hline
\multicolumn{1}{|l|}{res\_4}     & \multicolumn{1}{l|}{256}    & \multicolumn{1}{l|}{256}      & \multicolumn{1}{l|}{12}      & \multicolumn{1}{l|}{2}      & \multicolumn{1}{l|}{5}   & \multicolumn{1}{l|}{1/4}        & \multicolumn{1}{l|}{N/16}     \\ \hline
\multicolumn{1}{|l|}{res\_t\_5}     & \multicolumn{1}{l|}{256}    & \multicolumn{1}{l|}{256}     & \multicolumn{1}{l|}{8}      & \multicolumn{1}{l|}{2}      & \multicolumn{1}{l|}{3}   & \multicolumn{1}{l|}{1/5}        & \multicolumn{1}{l|}{N/8}     \\ \hline
\multicolumn{1}{|l|}{res\_t\_6}     & \multicolumn{1}{l|}{256}   & \multicolumn{1}{l|}{256}     & \multicolumn{1}{l|}{8}      & \multicolumn{1}{l|}{2}      & \multicolumn{1}{l|}{3}   & \multicolumn{1}{l|}{1/6}        & \multicolumn{1}{l|}{N/4}     \\ \hline
\multicolumn{1}{|l|}{res\_t\_7}    & \multicolumn{1}{l|}{256}   & \multicolumn{1}{l|}{128}        & \multicolumn{1}{l|}{8}      & \multicolumn{1}{l|}{2}      & \multicolumn{1}{l|}{3}   & \multicolumn{1}{l|}{1/7}           & \multicolumn{1}{l|}{N/2}     \\ \hline
\multicolumn{1}{|l|}{res\_t\_8}    & \multicolumn{1}{l|}{128}   & \multicolumn{1}{l|}{128}        & \multicolumn{1}{l|}{8}      & \multicolumn{1}{l|}{2}      & \multicolumn{1}{l|}{3}   & \multicolumn{1}{l|}{1/8}           & \multicolumn{1}{l|}{N}     \\ \hline
\multicolumn{1}{|l|}{res\_9}    & \multicolumn{1}{l|}{128}   & \multicolumn{1}{l|}{3}        & \multicolumn{1}{l|}{5}      & \multicolumn{1}{l|}{1}      & \multicolumn{1}{l|}{2}   & \multicolumn{1}{l|}{1/9}           & \multicolumn{1}{l|}{N}     \\ \hline
\end{tabular}
\caption{Network Architectures. res indicates the 1D residual convolution layer. res\_t refers to the 1-D transpose residual convolution layer. in\_ch is the input channel number, out\_ch is the output channel number, ratio is the residual ratio defined in \sect{sec:generation} and out\_len is the output sequence length of each layer. }
\label{tab:network_architectures}
\end{table*}

%% file: table/rotation_range.tex
\begin{table*}[]
\begin{tabular}{|l|l|l|l|l|l|l|l|l|l|}
\hline
Joint         & x range   & y range   & z range   & Order & Joint        & x range   & y range  & z range  & Order \\ \hline
hip           & -360, 360 & -360, 360 & -360, 360 & xzy   & lCollar      & -30, 30   & -30, 30  & -10, 10  & xzy   \\ \hline
abdomen       & -45, 68   & -30, 30   & -45, 45   & xzy   & lShldr       & -90, 135  & -110, 30 & -70, 60  & xzy   \\ \hline
chest         & -45, 45   & -30, 30   & -45, 45   & xzy   & lForeArm     & 0, 0      & 0, 150   & -30, 120 & yxz   \\ \hline
neck          & -37, 22   & -30, 30   & -45, 45   & xzy   & lHand        & -90, 90   & -20, 30  & 0, 0     & xzy   \\ \hline
head          & -37, 22   & -30, 30   & -45, 45   & xzy   & lThumb1      & 0, 0      & 0, 0     & 0, 0     & xzy   \\ \hline
leftEye       & 0, 0      & 0, 0      & 0, 0      & xzy   & lThumb2      & 0, 0      & 0, 0     & 0, 0     & xzy   \\ \hline
leftEye\_Nub  & -         & -         & -         & xzy   & lThumb\_Nub  & -         & -        & -        & xzy   \\ \hline
rightEye      & 0, 0      & 0, 0      & 0, 0      & xzy   & lIndex1      & 0, 0      & 0, 0     & 0, 0     & xzy   \\ \hline
rightEye\_Nub & -         & -         & -         & xzy   & lIndex2      & 0, 0      & 0, 0     & 0, 0     & xzy   \\ \hline
rCollar       & -30, 30   & -30, 30   & -10, 10   & xzy   & lIndex2\_Nub & -         & -        & -        & xzy   \\ \hline
rShldr        & -90, 135  & -30, 110  & -70, 60   & xzy   & lMid1        & 0, 0      & 0, 0     & 0, 0     & xzy   \\ \hline
rForeArm      & 0, 0      & 0, 150    & -30, 120  & yxz   & lMid2        & 0, 0      & 0, 0     & 0, 0     & yxz   \\ \hline
rHand         & -90, 90   & -30, 20   & 0, 0      & xzy   & lMid\_Nub    & -         & -        & -        & xzy   \\ \hline
rThumb1       & 0, 0      & 0, 0      & 0, 0      & xzy   & lRing1       & 0, 0      & 0, 0     & 0, 0     & xzy   \\ \hline
rThumb2       & 0, 0      & 0, 0      & 0, 0      & xzy   & lRing2       & 0, 0      & 0, 0     & 0, 0     & xzy   \\ \hline
rThumb\_Nub   & -         & -         & -         & xzy   & lRing\_Nub   & -         & -        & -        & xzy   \\ \hline
rIndex1       & 0, 0      & 0, 0      & 0, 0      & xzy   & lPinky1      & 0, 0      & 0, 0     & 0, 0     & xzy   \\ \hline
rIndex2       & 0, 0      & 0, 0      & 0, 0      & xzy   & lPinky2      & 0, 0      & 0, 0     & 0, 0     & xzy   \\ \hline
rIndex2\_Nub  & -         & -         & -         & xzy   & lPinky\_Nub  & -         & -        & -        & xzy   \\ \hline
rMid1         & 0, 0      & 0, 0      & 0, 0      & xzy   & rButtock     & -20, 20   & -20, 20  & -10, 10  & xzy   \\ \hline
rMid2         & 0, 0      & 0, 0      & 0, 0      & xzy   & rThigh       & -180, 100 & -180, 90 & -60, 70  & xzy   \\ \hline
rMid\_Nub     & -         & -         & -         & xzy   & rShin        & 0, 170    & 0, 0     & 0, 0     & xzy   \\ \hline
rRing1        & 0, 0      & 0, 0      & 0, 0      & xzy   & rFoot        & -31, 63   & -30, 30  & -20, 20  & xzy   \\ \hline
rRing2        & 0, 0      & 0, 0      & 0, 0      & xzy   & rFoot\_Nub   & -         & -        & -        & xzy   \\ \hline
rRing\_Nub    & -         & -         & -         & xzy   & lButtock     & -20, 20   & -20, 20  & -10, 10  & xzy   \\ \hline
rPinky1       & 0, 0      & 0, 0      & 0, 0      & xzy   & lThigh       & -180, 100 & -180, 90 & -60, 70  & xzy   \\ \hline
rPinky2       & 0, 0      & 0, 0      & 0, 0      & xzy   & lShin        & 0, 170    & 0, 0     & 0, 0     & xzy   \\ \hline
rPinky\_Nub   & -         & -         & -         & xzy   & lFoot        & -31, 63   & -30, 30  & -20, 20  & xzy   \\ \hline
              &           &           &           & xzy   & lFoot\_Nub   & -         & -        & -        & xzy   \\ \hline
\end{tabular}
\caption{Range and order of the rotations at each joint defined in the CMU Motion Capture Dataset. Note that finger motions are not captured in this dataset.}
\label{tab:rotation}
\end{table*}